\pdfoutput=1
\PassOptionsToPackage{table,dvipsnames}{xcolor}
\documentclass[11pt]{article}

\usepackage[final]{acl}

\usepackage{times}
\usepackage{latexsym}

\usepackage[T1]{fontenc}
\usepackage[utf8]{inputenc}

\usepackage{microtype}

\usepackage{inconsolata}

\usepackage{graphicx}

\usepackage{enumitem}
\usepackage{svg}
\usepackage{booktabs}

\usepackage{colortbl}
\usepackage{pgfmath}
\usepackage[most]{tcolorbox}

\usepackage{url}

\usepackage[most]{tcolorbox} 
\tcbset{
  myboxstyle/.style={
    colback=gray!10!white,   % Light gray background
    colframe=gray!70!black,  % Dark gray border
    fonttitle=\bfseries,     % Bold title
    coltitle=white,          % Title text color
    boxrule=0.3mm,           % Thinner border
    arc=2mm,                 % Slightly rounded corners
    left=1mm, right=1mm,     % Horizontal padding
    top=1mm, bottom=1mm      % Vertical padding
  }
}

\usepackage{siunitx}         % for number printing
\usepackage{pgf}             % for pgfmath
\usepackage{ifthen}          % for conditionals (if needed)

\usepackage{multirow}

\newcommand{\perc}[2]{%
  \pgfmathsetmacro{\increase}{((#2 - #1) / #1 * 100)}%
  \pgfmathparse{round(\increase)}%
  \pgfmathtruncatemacro{\increaseInt}{\pgfmathresult}%
  \pgfmathtruncatemacro{\displayPerc}{abs(\increaseInt)}%
  \pgfmathsetmacro{\colorlevel}{max(0, min(100, \displayPerc * 3))}%
  \ifdim \increase pt > 0pt
    \def\arrowSymbol{$\uparrow$}%
    \def\cellColor{green}%
  \else
    \def\arrowSymbol{$\downarrow$}%
    \def\cellColor{red}%
  \fi
  \edef\temp{\noexpand\cellcolor{\cellColor!\colorlevel} \num[round-precision=0]{#2}\, \arrowSymbol\displayPerc\%\ }%
  \temp%
}

\newcommand{\percsecnew}[2]{%
  \pgfmathsetmacro{\increase}{((#2 - #1) / #1 * 100)}%
  \pgfmathparse{round(\increase)}%
  \pgfmathtruncatemacro{\increaseInt}{\pgfmathresult}%
  \pgfmathtruncatemacro{\displayPerc}{abs(\increaseInt)}%
  \pgfmathsetmacro{\colorlevel}{max(0, min(100, \displayPerc * 3))}%
  \ifdim \increase pt > 0pt
    \def\arrowSymbol{$\uparrow$}%
    \def\cellColor{green}%
  \else
    \def\arrowSymbol{$\downarrow$}%
    \def\cellColor{red}%
  \fi
  \edef\temp{\noexpand\cellcolor{\cellColor!\colorlevel} \arrowSymbol\displayPerc\%\, \num[round-precision=0]{#2}  }%
  \temp%
}

\newcommand{\percnew}[2]{%
  \pgfmathsetmacro{\increase}{((#2 - #1) / #1 * 100)}%
  \pgfmathtruncatemacro{\increaseInt}{\increase}%
  \pgfmathtruncatemacro{\displayPerc}{abs(\increaseInt)}%
  \ifdim \increase pt > 0pt
  \pgfmathsetmacro{\colorlevel}{max(0, min(100, \displayPerc * 3))}%
    \def\arrowSymbol{$\uparrow$}%
    \def\cellColor{green}%
  \else
    \pgfmathsetmacro{\colorlevel}{max(0, min(100, \displayPerc * 1))}%
    \def\arrowSymbol{$\downarrow$}%
    \def\cellColor{red}%
  \fi
  \edef\temp{\noexpand\cellcolor{\cellColor!\colorlevel} \num[round-precision=0]{#2}\,\arrowSymbol }%
  \temp%
}

\title{PCoT: Persuasion-Augmented Chain of Thought for Detecting Fake News and Social Media Disinformation}

\author{
 \textbf{Arkadiusz Modzelewski\textsuperscript{1,2}},
 \textbf{Witold Sosnowski\textsuperscript{2}},
 \textbf{Tiziano Labruna\textsuperscript{1}}, \\
 \textbf{Adam Wierzbicki\textsuperscript{2}},
 \textbf{Giovanni Da San Martino\textsuperscript{1}}
\\
\\
 \textsuperscript{1}University of Padua, Italy \\
 \textsuperscript{2}Polish-Japanese Academy of Information Technology, Poland
\\
 \small{
   \textbf{Correspondence:} \href{mailto:contact@amodzelewski.com}{contact@amodzelewski.com}
 }
}

\begin{document}
\maketitle
\begin{abstract}

Disinformation detection is a key aspect of media literacy. Psychological studies have shown that knowledge of persuasive fallacies helps individuals detect disinformation. Inspired by these findings, we experimented with large language models (LLMs) to test whether infusing persuasion knowledge enhances disinformation detection. As a result, we introduce the Persuasion-Augmented Chain of Thought (PCoT), a novel approach that leverages persuasion to improve disinformation detection in zero-shot classification. We extensively evaluate PCoT on online news and social media posts. Moreover, we publish two novel, up-to-date disinformation datasets: EUDisinfo and MultiDis. These datasets enable the evaluation of PCoT on content entirely unseen by the LLMs used in our experiments, as the content was published after the models' knowledge cutoffs. We show that, on average, PCoT outperforms competitive methods by 15\% across five LLMs and five datasets. These findings highlight the value of persuasion in strengthening zero-shot disinformation detection.

\end{abstract}

\section{Introduction}
The spread of disinformation in digital communication poses significant risks to the state of democracy by shaping public opinion, reinforcing ideological divides, and fostering distrust in political institutions. \cite{jungherr2024negative, farhall2019political, brummette2018read}. The growing accessibility of digital media, coupled with reduced funds for traditional fact-checking efforts and the rise of alternatives like Birdwatch on Platform X (formerly Twitter), underscores the urgent need for complementary disinformation detection systems \cite{saeed2022crowdsourced, allen2022birds}. Traditional supervised detection methods, which rely on human-annotated data, face challenges in generalization and the scarcity of labeled data. This reinforces the need for zero-shot detection systems. 

\begin{figure}[ht]
    \centering
    \includegraphics[width=0.49\textwidth]{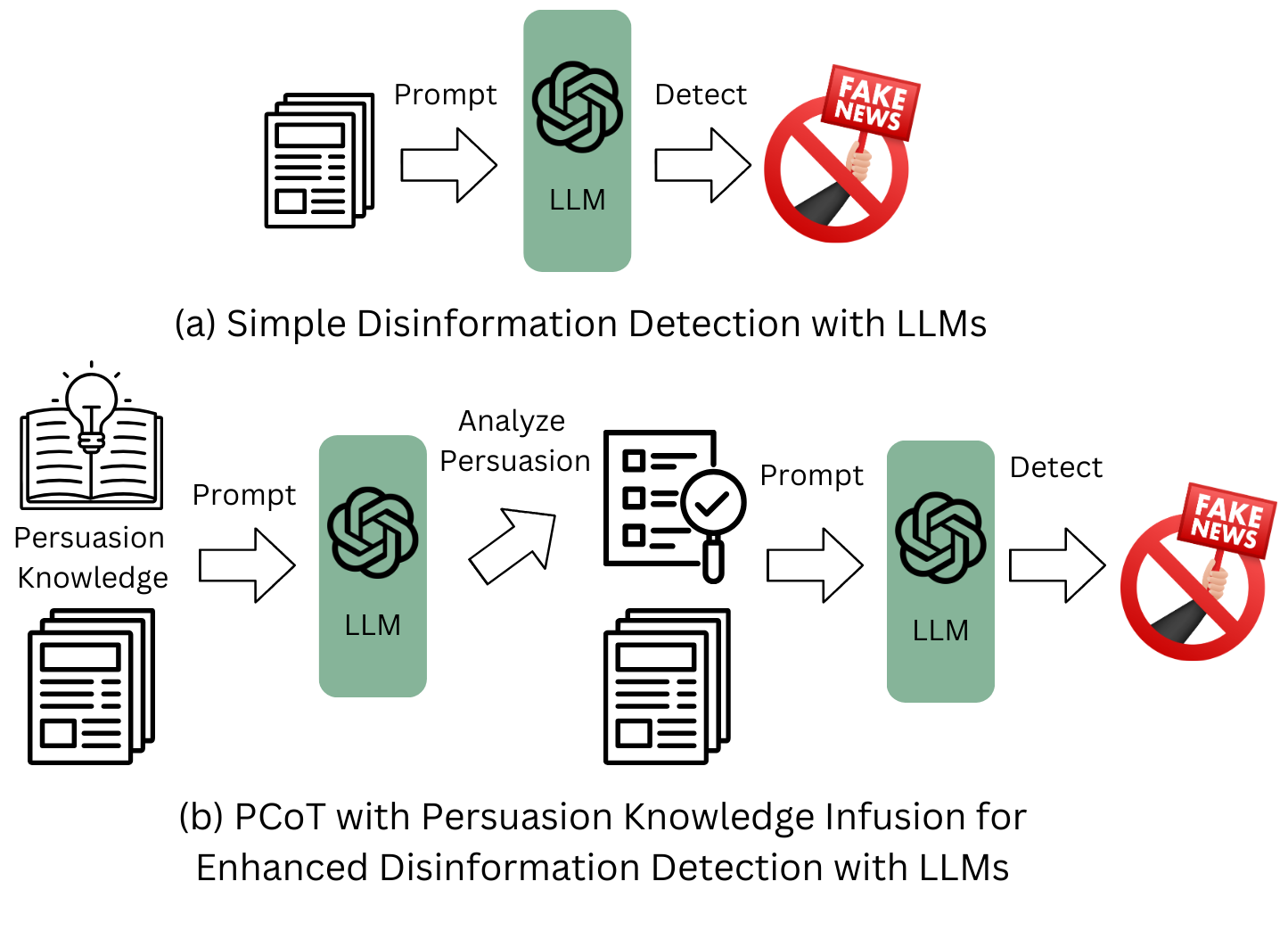}
    \caption{The comparison between detecting disinformation with LLMs in a simple zero shot setting and detecting with PCoT and infused knowledge about persuasion.}
    \label{fig:pcot_vs_simple}
\end{figure}

A critical aspect of disinformation is its coexistence with manipulation and persuasion to mislead audiences \cite{modzelewski2024mipd, chen2021persuasion}. 
Psychological studies show that teaching individuals to recognize persuasive fallacies improves their ability to distinguish between real and fake news \cite{hruschka2023learning}. Building on this, we explored whether infusing knowledge of persuasion into generative LLMs enhances disinformation detection. 

%we introduce a novel zero-shot method that more effectively addresses generalization and data scarcity challenges compared to supervised models.
As a result, we present \textbf{Persuasion-Augmented Chain of Thought (PCoT)}, a novel zero-shot method leveraging persuasion signals to improve disinformation detection that more effectively addresses generalization and annotated data scarcity challenges compared to supervised models.
% Recent psychological research has shown that teaching individuals to recognize persuasive logical fallacies can significantly improve their ability to discern between real and fake news \cite{hruschka2023learning}. Building on these findings, we explored whether incorporating knowledge of persuasion strategies into generative large language models (LLMs) could enhance their ability to detect disinformation. As a result, we introduce a novel zero-shot method that more effectively addresses generalization and data scarcity challenges compared to supervised models.
% As a result, we propose a novel approach that explicitly incorporates the analysis of persuasion strategies to improve detection performance in zero-shot settings.
%This paper introduces \textbf{Persuasion-Augmented Chain of Thought (PCoT)}, a novel method for leveraging persuasion signals to improve disinformation detection using generative LLMs in zero-shot settings. 
PCoT operates through a two-stage process where the LLM first identifies and analyzes persuasion within a given text, using infused knowledge. This analysis is then utilized in subsequent reasoning to determine the presence of disinformation. By augmenting models's decision-making process with persusion knowledge, PCoT achieves significant gains in detection performance across multiple datasets. 

We conducted experiments on five datasets covering fake news and social media disinformation to evaluate our method rigorously. We evaluated PCoT on two novel datasets, MultiDis and EUDisinfo, which contain up-to-date articles from 2024 onwards, ensuring they were not part of the pretraining data for any tested LLMs. The \textbf{Multi}topic \textbf{Dis}information is a high-quality dataset developed with fact-checking experts with prior experience in debunking organizations accredited by the International Fact-Checking Network\footnote{The International Fact-Checking Network gives accreditation to debunking and fact-checking organizations that sign its code of principles. See \href{https://www.poynter.org/ifcn/}{https://www.poynter.org/ifcn/}}. For a comprehensive evaluation, we also used three publicly available datasets containing texts before the knowledge cutoff of all tested models.

For evaluation, we selected three top-performing methods on zero-shot disinformation detection from \citet{lucas2023fighting} and adapted them to incorporate our PCoT approach. Using five different LLMs, we demonstrate that PCoT delivers significant performance improvements over chosen competitive methods.

Our main contributions are as follows:

\begin{itemize}[nosep, leftmargin=*]
    \item We introduce a novel Persuasion-Augmented Chain of Thought (PCoT) method that significantly enhances the ability of generative LLMs to detect disinformation in a zero-shot setting.
    \item We provide a thorough analysis of the effectiveness of our method across various datasets, including fake news and disinformative social media posts from X (formerly Twitter).
    \item  We introduce two novel datasets, MultiDis and EUDisinfo, to further assess our method. These datasets consist of articles collected after the knowledge cutoff of the tested LLMs, allowing for a thorough and rigorous PCoT evaluation on texts that were entirely unseen by the models. 
    \item We analyze how LLM-predicted persuasion affects disinformation detection effectiveness.
\end{itemize}
We release the final prompts and the codebase\footnote{Repository with data, prompts and codebase: \url{https://github.com/ArkadiusDS/PCoT}}.

\section{Datasets}
\label{sec:dataset}

To ensure robust performance of our method across diverse data conditions and inspired by the work of \citet{lucas2023fighting}, we designed our evaluation to address potential dataset overlap with LLM pretraining. We tested our method on two dataset types: (i) prior-cutoff datasets, which may contain pretraining content, and (ii) two novel datasets of articles published after the models' knowledge cutoff. This setup enables a rigorous evaluation of our PCoT method on potential pretraining content and entirely new information. Moreover, we evaluated our method on social media posts versus longer articles, such as news.

\subsection{Prior-Cutoff Datasets}  
The following datasets, published before January 1, 2024, may overlap with the models' training data.  

\begin{itemize}[nosep, leftmargin=*]  
    \item \textbf{CoAID} – A dataset for COVID-19 misinformation detection, comprising 4k+ news articles and 1k+ social posts, all annotated with ground-truth labels \cite{cui2020coaid}.  
    \item \textbf{ISOT Fake News} – A dataset of 44k+ fake and truthful articles from reputable and unreliable sources, identified via Politifact\footnote{PolitiFact is a nonprofit fact-checking project by the Poynter Institute.} \cite{ahmed2018detecting, ahmed2017detection}.  
    \item \textbf{ECTF} – An extended version of CTF \cite{paka2021cross} for detecting fake news on Platform X about COVID-19, with additional data to improve early-stage detection \cite{bansal2021combining}.  
\end{itemize}

\subsection{Novel Post-Cutoff Datasets}

\subsubsection{MultiDis}  
The \textbf{Multi}topic \textbf{Dis}information Dataset comprises nearly 2,000 English articles on European and global disinformation. 
%Created 
It has been created by researchers from multiple European universities to support disinformation detection research.  

\paragraph{Annotation Process}
The annotation process involved four key stages:  
\begin{enumerate}[nosep, leftmargin=*]  
    \item \textbf{Methodology and Data Preparation} – Researchers, fact-checking and debunking experts developed a robust methodology and guidelines before collecting a database of articles.  
    \item \textbf{In-Depth Training} – A three-day hybrid training led by the most experienced fact-checking expert aimed to deliver in-depth on-site training to all European teams while ensuring accessibility for remote annotators. Each team was assigned two supervisors, usually a disinformation researcher. The training concluded with an initial annotation round, reviewed and discussed by a fact-checking expert. These preliminary annotations were excluded from the final dataset to maintain high quality.
    \item \textbf{Article Annotation} – Independent annotation by a less experienced annotator and a supervisor.  
    \item \textbf{Final Evaluation} – The supervisor reviewed both annotations and resolved disagreements through discussion when necessary. A senior fact-checking expert contributed when needed. If consensus was unattainable, the article was labeled \textit{Hard-to-say}.  
\end{enumerate} 
Appendix \ref{sec:methodology} shows dataset and annotation details.  
 
\paragraph{Data Sources}  

We selected diverse sources to ensure access to both reliable and unreliable content, categorizing each as \textit{Reliable}, \textit{Unreliable}, or \textit{Mixed}. A team of experts evaluated sources through consensus, thoroughly analyzing the source's regularly published content and cross-checking with established tools (e.g., Media Bias/Fact Check). The assigned categories were not revealed to annotators to prevent biases toward sources.

The MultiDis dataset includes a variety of sources: global news agencies, regional publications, thematic platforms, fact-checking organizations, and independent media. All used 44 distinct sources are freely accessible. To ensure transparency, we make these sources publicly available.

\paragraph{Thematic Category} Before detailed analysis, articles were manually assigned to one of eight thematic categories. The selection of these topics was informed by the EU DisinfoLab report\footnote{\href{https://www.disinfo.eu/publications/connecting-the-disinformation-dots/}{The EU DisinfoLab’s report}, grounded in expert research from 20 countries across Europe, guarantees high quality and credibility.} \cite{kn:Sessa}. The categories are: (i) \textit{Anti-Europeanism and Anti-Atlanticism}; (ii) \textit{Anti-migration and Xenophobia}; (iii) \textit{Climate Change and the Energy Crisis}; (iv) \textit{Health}; (v) \textit{Institutional and Media Distrust}; (vi) \textit{Gender Issues}; (vii) \textit{Ukraine War and Refugees}; (viii) \textit{LGBT+}. Table \ref{tab:articles_per_category} shows the distribution of articles by thematic category in the MultiDis dataset.

Annotators, during the credibility evaluation, could label articles as \textit{Inconsistent with the topic}, excluding them from further analysis to ensure high-quality topic assignments.

\begin{table}[ht]
    \centering
    \renewcommand{\arraystretch}{1.2} % Increase vertical spacing for readability
\setlength{\tabcolsep}{5pt} % Reduce horizontal spacing between columns
   % \scriptsize
   \footnotesize
    \begin{tabular}{l c c}
        \toprule
        \textbf{Category} & \textbf{\textit{\#DOC}}  & \textbf{\textit{\#PERC}}\\
        \midrule
        Anti-Europeanism \& Anti-Atlanticism & 219  & 11.4\% \\
        Anti-migration and Xenophobia & 117  & 6.1\% \\
        Climate Change and the Energy Crisis & 324  & 16.9\% \\
        Health & 285  & 14.8\% \\
        Institutional and Media Distrust & 317  & 16.5\%\\
        Gender Issues & 97  & 5.0\%\\
        Ukraine War and Refugees & 361  &  18.8\%\\
        LGBT+ & 202  & 10.5\%\\ 
        % \bottomrule
        % Total & 1922  &  \\
        \bottomrule
    \end{tabular}
    \caption{Number of articles (\textit{\#DOC}) per thematic category and their (\textit{\#PERC}) percentage in MulitDis dataset.}
    \label{tab:articles_per_category}
\end{table}

\paragraph{Credibility Analysis}
Annotators assessed each article using a debunking technique, auxiliary complemented by fact-checking, as defined by the NATO Strategic Communications Centre of Excellence \cite{pamment2021fact}.

Given an article, annotators analyze its content to determine whether it
belongs to one of four categories. The main categories in our guidelines are: \textit{Credible Information}, \textit{Disinformation}, with the latter following the European Commission's High-Level Expert Group: \textit{Disinformation is false, inaccurate or misleading information designed, presented, and promoted to intentionally cause public harm or for profit} \cite{de2018multi}. This definition has also been adopted in other disinformation studies \cite{modzelewski2024mipd, sosnowski2024eu}. Two additional labels, \textit{Hard-to-say} and \textit{Inconsistent with the Topic}, were respectively assigned to articles where annotators did not reach a consensus or where the content did not match the assigned topic. Articles labeled with these or published before January 2024 were excluded from experiments.

\paragraph{Bias Prevention and Data Quality}
Our guidelines require each article to be annotated independently by two experts to minimize bias. Given the time demands of the annotation process, only two independent evaluations per article were guaranteed. 
Supervisors provided the third final annotation by reviewing the two previous annotations. However, supervisors were instructed to resolve uncertainties through discussions, and the lead fact-checking expert provided clarification when needed. These discussions helped ensure consistency among annotators and reduced human errors and bias. We achieved full agreement in the first two rounds for 86.78\% of articles, with the remainder undergoing a more detailed third analysis.

\noindent\textbf{Note}: We publicly release the complete dataset, including annotations from all three rounds.

\subsubsection{EUDisinfo}
We introduce the EUDisinfo dataset, collected with usage of the EUvsDisinfo database\footnote{\url{https://EUvsDisinfo.eu/disinformation-cases/}}, which comprises 18,464 disinformation cases\footnote{Database size recorded as of February 11, 2025.}. EUvsDisinfo is an EU initiative dedicated to identifying, analyzing, and countering pro-Kremlin disinformation. Each entry concisely summarizes a disinformation case, along with links to the original misleading content and credible sources debunking the claims. Since EUvsDisinfo provides predefined %credibility 
evaluation for each disinformation case as either \textit{credible} or \textit{disinformation}, we did not conduct additional annotation. 
% Each collected article retains the original classification based on EUvsDisinfo’s assessment. 
The EUvsDisinfo database comprises articles published in multiple languages, some previously analyzed in \citet{leite2024euvsdisinfo}. However, as all articles in that study date before 2024, this dataset was unsuitable for our research. To address this limitation, we independently curated a collection of approximately 400 English articles published in 2024 or later.

\begin{table}[h!]
%\scriptsize
\footnotesize
\centering
\begin{tabular}{l|c|c}
\toprule
\textbf{Category} & \textbf{MultiDis} & \textbf{EUDisinfo} \\
\midrule
Credible Information & 65.3\% & 67.1\% \\
Disinformation        & 32.8\% & 32.9\% \\
\bottomrule
\end{tabular}

\caption{Percentage of articles per main credibility category in MultiDis and EUDisinfo datasets.}
\label{tab:articles_per_category_combined}
\end{table}

To collect English news article content, we leveraged the \textit{Trafilatura} tool~\cite{barbaresi-2021-trafilatura}, which efficiently scrapes web content while preserving article structure. Additionally, we employed \textit{Selenium} \cite{selenium} to navigate and extract HTML pages and \textit{Beautiful Soup 4} \cite{beautifulsoup4} to parse article content.

Table \ref{tab:articles_per_category_combined} presents the percentage of articles per main credibility category in our two datasets. More detailed statistics are provided in Appendix \ref{sec:app_data_stats}.

\section{Proposed Method}
In this section, we introduce the Persuasion-Augmented Chain of Thought (PCoT) method, which leverages persuasion to enhance zero-shot disinformation detection using generative LLMs. 

\subsection{Persuasion-Augmented Chain of Thought}
Empirical studies have shown that persuasion is an integral part of disinformation 
  \cite{chen2021persuasion}. Insights from psychological research highlight the potential of leveraging persuasion knowledge to more effectively discern between fake and credible news \cite{hruschka2023learning}. 
  % Persuasive strategies employed in disinformation can hinder its detection, posing a difficulty even for adult humans \cite{peng2024navigating}. 
  Inspired by this, we propose the Persuasion-Augmented Chain of Thought. The PCoT method employs a two-stage reasoning process that improves LLM's disinformation detection by persuasion knowledge infusion.

In the first stage, an LLM is prompted to perform multi-faceted reasoning by analyzing persuasion strategies (see Figure \ref{fig:persuasion_strategies_tax}) within the text. The second stage performs the disinformation detection task, enriched by the previously generated analysis of persuasion strategies. Figure \ref{fig:pcot_vs_simple} presents a simplified comparison between traditional zero-shot disinformation detection using LLMs and our PCoT method. Final prompt templates for each stage of our PCoT method are available in Appendix \ref{sec:prompts} .

\subsection{Persuasion Detection Step}

In the first stage LLM performs multifaceted reasoning by tackling the multi-class, multi-label task of detecting persuasion strategies, along with contextual question answering by explaining persuasion usage within each text. The persuasion detection task can be formally represented as follows: 
The model \( M \) takes as input the text \( T \), the impersonation \( I_{P} \), the infused knowledge \( K_P \) and guidelines \( G_P \). Here, \( I_P \) establishes the context and overrides alignment tuning, while \( K_P \) encapsulates knowledge about a predefined set of high-level persuasion strategies \( P \), and guidelines \( G_P \) that determine the task and specify the structure of the expected response. This combined input is represented as \( X = (T, I_{P}, K_P, G_P) \), where the set of persuasion strategies is given by \( P = \{p_1, p_2, \ldots, p_k\} \). For each text, the model generates an output in a structured textual format that can be decoded into a \textit{JSON}-like dictionary. This output contains, for each persuasion strategy \( p_i \in P \), two components: a binary label \( y_{p_i} \) (\textit{‘Yes’} or \textit{‘No’}) indicating the presence of \( p_i \) in the text, and an explanation \( E_{p_i} \) justifying the prediction. The output can be formally expressed as:
\begin{equation}
  \label{eq:per_dict}
A_T = \{p_i : (y_{p_i}, E_{p_i}) \mid p_i \in P\}.
\end{equation}

The model  \( M \) generates the output \( A_T \) by leveraging the combined input \( X \), capturing both the text and infused persuasion knowledge:
\begin{equation}
  \label{eq:per_pred}
 A_T \sim M(T, I_{P}, K_P, G_P).
\end{equation}

 This stage leverages the capabilities of generative LLMs to integrate knowledge about persuasion into the reasoning process. 
 The rationale for our approach is based on the observations that explanations can enhance the robusteness of the final prediction~\cite{he2024using}, and that previous works have shown
% In PCoT, we use LLM-generated 
%explanations to enhance the robustness of the final results \cite{he2024using}. Furthermore, research has shown 
that incorporating explanations can improve zero-shot classification performance \cite{menon2022clues}.

\subsection{Disinformation Detection Step}

In the final stage of the PCoT method, LLM performs zero-shot binary classification on each input text. Formally, the model \( M \) evaluates the input text \( T \) to detect disinformation. It processes the combined input \( X = (T, I_{D}, A_T, G_D) \), where \( I_{D} \) defines the impersonation that establishes the context, \( A_T \) provides the persuasion analysis from the first stage of PCoT, and \( G_D \) defines the task and specifies the structure of the expected response. The model then generates the output, \( Y_T \) indicating whether \( T \) contains disinformation (\textit{‘Yes’} or \textit{‘No’}).  

\begin{equation}
  \label{eq:example}
Y_T  \sim M(T, I_{D}, A_T, G_D).
\end{equation}

 We explored the zero-shot setting as many studies have shown that zero-shot prompting of LLMs like GPT-4 can outperform supervised models like BERT in detecting disinformation \cite{pelrine2023towards, bang2023multitask, hassan2020political}. In addition, \citet{lucas2023fighting} demonstrated that fine-tuning BERT on different datasets and testing on unseen data leads to worse performance than zero-shot with LLMs. We confirm these findings on our data, as outlined in the Appendix \ref{sec:bert_vs_llm}.

\section{Experiments and Evaluation}
\label{sec:experiments}
We created five test sets by randomly selecting texts from each dataset. To evaluate our PCoT method’s ability to detect disinformation in data unseen by LLMs, we used two novel test sets, MultiDis and EUDisinfo. These test sets contain only articles published from 2024 onward, ensuring that the content was not part of any LLM training data. Each test set contained 400-500 articles or posts. Appendix \ref{sec:test_data_stats} provides statistics on the composition of the test datasets used in our experiments.

We conducted all experiments on five different LLMs: \textit{GPT 4o Mini}, \textit{Llama 3.1 8B}, \textit{Claude 3 Haiku}, \textit{Llama 3.3 70B}, and \textit{Gemini 1.5 Flash}. To ensure the most deterministic results possible, we set the hyperparameter temperature to 0 in each model. Appendix \ref{sec:models} includes more details about the models used, including knowledge cutoff dates and the rationale behind our choice.

% We evaluated PCoT using the F\textsubscript{1} score. We used McNemar's test to assess the significance of the difference between PCoT and competitive methods, as it suits binary tasks comparing two methods on the same dataset \cite{dror2018hitchhiker, dietterich1998approximate}. It has been widely applied in NLP \cite{card2020little, blitzer2006domain}.

PCoT was evaluated using the F\textsubscript{1} score. To assess the significance of its difference from competitive methods, we used McNemar's test, which suits binary tasks comparing two methods on the same dataset \cite{dror2018hitchhiker, dietterich1998approximate}. This statistical test has been widely applied in NLP \cite{card2020little, blitzer2006domain}.

\begin{figure}[ht]
  \fbox{
    \begin{minipage}{0.94\columnwidth}
      \scriptsize % Makes all text smaller
      \colorbox{blue!20}{\textbf{Attack on reputation [AR]}} - the argument does not address the topic itself but targets the participant (personality, experience, etc.) to question and/or undermine their credibility. The object of the argumentation can also refer to a group of individuals, an organization, an object, or an activity.
      
      \colorbox{blue!20}{\textbf{Justification [J]}} - the argument is made of two parts, a statement and an explanation or appeal, where the latter is used to justify and/or to support the statement.
      
      \colorbox{blue!20}{\textbf{Simplification [S]}} - the argument excessively simplifies a problem, usually regarding the cause, the consequence, or the existence of choices.
      
      \colorbox{blue!20}{\textbf{Distraction [D]}} - the argument takes focus away from the main topic or argument to distract the reader.
      
      \colorbox{blue!20}{\textbf{Call [C]}} - the text is not an argument, but an encouragement to act or to think in a particular way.
      
      \colorbox{blue!20}{\textbf{Manipulative wording [MW]}} - the text is not an argument per se, but uses specific language, which contains words or phrases that are either non-neutral, confusing, exaggerating, loaded, etc., in order to impact the reader emotionally.
    \end{minipage}
  }
  \caption{Persuasion strategies used in our experiments. A detailed description of the techniques associated with these strategies can be found in Appendix \ref{sec:persuasion_strategies_techniques}.}
  \label{fig:persuasion_strategies_tax}
\end{figure}

\subsection{Persuasion Detection Step}

To enhance first stage of the PCoT method we created prompts with infused persuasion knowledge. The knowledge applied within the prompt is based on the taxonomy presented by \citet{piskorski2023news, piskorski2023multilingual}. This taxonomy categorizes persuasion techniques into six strategies (shortcuts in brackets): \textit{Attack on reputation [AR]}, \textit{Justification [J]}, \textit{Simplification [S]}, \textit{Distraction [D]}, \textit{Call [C]}, \textit{Manipulative wording [MW]} (definitions in Figure \ref{fig:persuasion_strategies_tax}). Using a well-established taxonomy enabled a thorough first-stage evaluation since datasets annotated with it are widely used \cite{dimitrov2024semeval, piskorski2023semeval}.

\begin{table}[ht]
    \centering
    \renewcommand{\arraystretch}{1.2}
\setlength{\tabcolsep}{5pt} % Reduce horizontal spacing between columns
    % \scriptsize % Make the font size smaller
    \footnotesize
    \begin{tabular}{l c}
        \toprule
         Method & F\textsubscript{1} Micro \\
        \midrule
        DMT &  \percsecnew{0.664}{0.722}$\pm0.035$ \\
        DTAT & \percsecnew{0.664}{0.689}$\pm0.042$  \\
        Base MT & 0.664$\pm0.030$ \\ 
                \bottomrule
    \end{tabular}
    \caption{Average F\textsubscript{1} micro ($\pm std$, over five LLMs) for three methods evaluated in the first stage of the PCoT method. Percentage changes are computed relative to the \textit{Base MT} method. The \textit{DMT} variant is selected as the final best-performing method for this stage.}
    \label{tab:persuasion_step_binary_vs_multi}
\end{table}

To develop the most effective prompt for detecting persuasive strategies, we conducted extensive experiments on the SemEval 2023 dataset \cite{piskorski2023semeval}, using 536 English news articles with ground truth on persuasion strategies and five LLMs. We used F\textsubscript{1} micro as the evaluation metric for this stage, following its use in a closely related task at SemEval 2023 \cite{piskorski2023semeval}. 

We tested various prompts, including: 
\begin{itemize}[nosep, leftmargin=*]
    \item Detailed Multitask (DMT) - a single prompt for detecting all strategies and their explanations. Prompt with infused knowledge about persuasion strategies and their definitions (see Figure~\ref{fig:persuasion_strategies_tax}), and the specific techniques with definitions (see Appendix \ref{sec:persuasion_strategies_techniques}) that fall under each strategy. These techniques are categorized according to the taxonomy proposed by \citet{piskorski2023news, piskorski2023multilingual}.
    \item Detailed One Task At a Time (DTAT) - individual prompts for binary detection and explanations per strategy, infusing the same knowledge as DMT but divided into six parts as there are six persuasion strategies.
    \item Base Multitask - our baseline single prompt for detecting all strategies. It does not incorporate persuasion knowledge but simply lists strategy names and prompts identification of those present in the text. This served as our starting point.
\end{itemize}
\noindent As shown in Table \ref{tab:persuasion_step_binary_vs_multi}, the DMT method achieved the highest F\textsubscript{1} micro score, outperforming our baseline prompt by 9\%. As a result, DMT was used in the first stage of our final PCoT method. Our experiments revealed important finding that:
\begin{tcolorbox}[
  colback=green!3!white, 
  colframe=green!40!black, 
  title= \textbf{\textcolor{white}{\small Finding}}, 
  boxsep=1pt, % Reduces padding inside the box
  left=1pt, % Reduces left margin
  right=1pt, % Reduces right margin
  top=1pt, % Reduces top padding
  bottom=1pt % Reduces bottom padding
]
\small Using a single prompt to identify all persuasion strategies was more effective than separate prompts for each strategy's binary classification.
\end{tcolorbox}

% The final, best-performing prompt used for the PCoT method includes persuasion strategies and their definitions (see Figure \ref{fig:persuasion_strategies_tax}), and the specific techniques with definitions that fall under each strategy. These techniques are categorized according to the taxonomy proposed by \citet{piskorski2023news, piskorski2023multilingual} and are listed in Appendix \ref{sec:persuasion_strategies_techniques}.

The persuasion detection step provides an analysis that includes binary labels and explanations. To assess the impact of these explanations, we also evaluated PCoT without them. Testing PCoT without explanations showed that including LLM-generated insights improved performance. 

This step establishes the foundation for the second stage of PCoT by analyzing the persuasion signals present in the input text. Appendix \ref{sec:persuasion_det_step_exp_and_impact_on_pcot} presents more details on tested prompts, evaluation results and rationale behind chosen taxonomy.

\subsection{Disinformation Detection Step}

For disinformation detection stage, we selected three top-performing competitive methods based on an extensive evaluation by \citet{lucas2023fighting}, specifically those that excelled on human-annotated datasets \cite{cui2020coaid, shu2020fakenewsnet}. We outline the three methods below:
\begin{itemize}[nosep, leftmargin=*] 
\item \textit{VaN} - A vanilla prompt serving as a fundamental baseline, offering concise instructions to LLMs \cite{lucas2023fighting}. 
\item \textit{Z-CoT} - Extends \textit{VaN} with a prompt encouraging step-by-step reasoning, inspired by \citet{kojima2022large}'s findings on zero-shot reasoning. 
\item \textit{DeF-SpeC} - Emphasizes contextual, deductive, and abductive reasoning \cite{lucas2023fighting}, addressing LLM limitations in inductive and multi-step reasoning \cite{bang2023multitask}. 
\end{itemize}
The chosen competitive methods served as baselines, allowing us to evaluate the effectiveness of PCoT. We then adapted these methods to our PCoT approach by modifying prompts to incorporate persuasion analysis from the first stage. This approach enabled us to determine whether the PCoT method is sensitive to prompt variations or exhibits consistent behavior. For a rigorous evaluation, we conducted experiments on five datasets covering various themes and genres, such as news and social media posts. The diverse selection of datasets allows us to assess PCoT's generalizability.

\begin{table}[ht]
    \centering
    \renewcommand{\arraystretch}{1.2}
\setlength{\tabcolsep}{5pt} % Reduce horizontal spacing between columns
    % \scriptsize % Make the font size smaller
    \footnotesize
    \begin{tabular}{l c}
        \toprule
         Method & F\textsubscript{1} Score \\
        \midrule
        PCoT &  \percsecnew{0.711}{0.815}$\pm0.027$ \\
        PCoT Single Step & \percsecnew{0.711}{0.765}$\pm0.072$ \\
        Base & 0.711$\pm0.055$ \\
                \bottomrule
    \end{tabular}
    \caption{Average F\textsubscript{1} ($\pm std$, over five LLMs) for \textit{PCoT} (two-stage) and \textit{PCoT Single Step}, which uses one prompt for simultaneous persuasion analysis and disinformation detection. Percentage changes are computed relative to the \textit{Base} method.}
    \label{tab:single_step}
\end{table}

\begin{table*}[!ht]
\scriptsize
\centering
\begin{tabular}{lcccccccccc}
\toprule
 & \multicolumn{2}{c}{\textbf{Overall}} & \multicolumn{2}{c}{\textbf{Articles}} & \multicolumn{2}{c}{\textbf{Posts}} & \multicolumn{2}{c}{\textbf{Prior Cutoff}} & \multicolumn{2}{c}{\textbf{Post Cutoff}} \\
\cmidrule(lr){2-11}
 & Base & PCoT & Base & PCoT & Base & PCoT & Base & PCoT & Base & PCoT \\
\midrule
\multicolumn{11}{l}{\textit{GPT 4o Mini}}\\
\quad VaN & 0.759 & \perc{0.759}{0.845} & 0.788 & \perc{0.788}{0.885}  & 0.700 & \perc{0.700}{0.762}  & 0.742 & \perc{0.742}{0.830}  & 0.790 & \perc{0.790}{0.874}  \\
\quad Z-CoT & 0.765 & \perc{0.765}{0.846}  & 0.801 & \perc{0.801}{0.884} & 0.696 & \perc{0.696}{0.767}  & 0.747 & \perc{0.747}{0.835}  & 0.801 & \perc{0.801}{0.869} \\
\quad DeF-SpeC & 0.772 & \perc{0.772}{0.834}  & 0.816 & \perc{0.816}{0.867}  & 0.690 & \perc{0.690}{0.766}  & 0.742 & \perc{0.742}{0.813}  & 0.832 & \perc{0.832}{0.875} \\
\midrule
\multicolumn{11}{l}{\textit{Gemini 1.5 Flash}}\\
\quad VaN & 0.681 & \perc{0.681}{0.810}  & 0.673 & \perc{0.673}{0.843} & 0.695 & \perc{0.695}{0.748}  & 0.683 & \perc{0.683}{0.778}  & 0.679 & \perc{0.679}{0.875} \\
\quad Z-CoT & 0.689 & \perc{0.689}{0.808}  & 0.681 & \perc{0.681}{0.838} & 0.703 & \perc{0.703}{0.752}  & 0.670 & \perc{0.670}{0.777}  & 0.687 & \perc{0.687}{0.872} \\
\quad DeF-SpeC & 0.744 & \perc{0.744}{0.834}  & 0.764 & \perc{0.764}{0.876} & 0.708 & \perc{0.708}{0.754}  & 0.721 & \perc{0.721}{0.810}  & 0.790 & \perc{0.790}{0.884} \\
\midrule
\multicolumn{11}{l}{\textit{Claude 3 Haiku}}\\
\quad VaN & 0.710 & \perc{0.710}{0.797}  & 0.714 & \perc{0.714}{0.820}  & 0.702 & \perc{0.702}{0.747}  & 0.728 & \perc{0.728}{0.797}  & 0.677 & \perc{0.677}{0.796} \\
\quad Z-CoT & 0.588 & \perc{0.588}{0.774}  & 0.601 & \perc{0.601}{0.800}  & 0.550 & \perc{0.550}{0.716}  & 0.565 & \perc{0.565}{0.767}  & 0.626 & \perc{0.626}{0.786} \\
\quad DeF-SpeC & 0.780 & \perc{0.780}{0.795}  & 0.806 & \perc{0.806}{0.810}  & 0.727 & \perc{0.727}{0.763}  & 0.809 & \perc{0.809}{0.812}  & 0.727 & \perc{0.727}{0.766} \\
\midrule
\multicolumn{11}{l}{\textit{Llama 3.3 70B }}\\
\quad VaN & 0.740 & \perc{0.740}{0.845}  & 0.747 & \perc{0.747}{0.881}  & 0.727 & \perc{0.727}{0.768}  & 0.733 & \perc{0.733}{0.839}  & 0.752 & \perc{0.752}{0.856} \\
\quad Z-CoT & 0.722 & \perc{0.722}{0.843}  & 0.725 & \perc{0.725}{0.878}  & 0.718 & \perc{0.718}{0.770}  & 0.707 & \perc{0.707}{0.837}  & 0.750 & \perc{0.750}{0.855} \\
\quad DeF-SpeC & 0.732 & \perc{0.732}{0.832}  & 0.740 & \perc{0.740}{0.863}  & 0.717 & \perc{0.717}{0.768}  & 0.719 & \perc{0.719}{0.806}  & 0.755 & \perc{0.755}{0.880} \\
\midrule
\multicolumn{11}{l}{\textit{Llama 3.1 8B}}\\
\quad VaN & 0.627 & \perc{0.627}{0.792}  & 0.565 & \perc{0.565}{0.802}  & 0.736 & \perc{0.736}{0.773}  & 0.649 & \perc{0.649}{0.788}  & 0.585 & \perc{0.585}{0.801}\\
\quad Z-CoT & 0.660 & \perc{0.660}{0.791}  & 0.623 & \perc{0.623}{0.804}  & 0.725 & \perc{0.725}{0.764}  & 0.670 & \perc{0.670}{0.789}  & 0.638 & \perc{0.638}{0.795} \\
\quad DeF-SpeC & 0.697 & \perc{0.697}{0.773}  & 0.688 & \perc{0.688}{0.784}  & 0.712 & \perc{0.712}{0.752}  & 0.683 & \perc{0.683}{0.767}  & 0.724 & \perc{0.724}{0.785} \\
\midrule
\textbf{Average} & 0.711 & \perc{0.711}{0.815} & 0.715 & \perc{0.715}{0.842} & 0.700 & \perc{0.700}{0.758} & 0.705 & \perc{0.705}{0.803} & 0.721 & \perc{0.721}{0.838} \\
\bottomrule
\end{tabular}
\caption{Results with F\textsubscript{1} scores  for five LLMs. The \textit{Base} columns shows the competitive method results, while the \textit{PCoT} columns presents results for prompts adapted to the PCoT method. McNemar's test confirmed that, across all models and methods, PCoT achieves significantly better results on \textit{Overall} data at the 0.01 significance level.}
\label{tab:f1_overall_scores}
\end{table*}

To demonstrate the need for two-stage PCoT, we tested a more straightforward single-step approach, where LLMs analyzed persuasion and detected disinformation simultaneously. As shown in Table \ref{tab:single_step} single-step PCoT outperformed the baseline by 8\%, while the two-stage method provided an additional significant 7\% improvement.

\section{Results and Discussion}
\paragraph{General Overview} The results of our experiments, presented in Table \ref{tab:f1_overall_scores}, compare the performance of our PCoT method with baseline approaches. PCoT significantly improves performance, achieving an average F\textsubscript{1} score of 0.815, about a 15\% improvement over the baselines. McNemar's test confirmed that, across all models and methods, PCoT consistently achieves significantly better results on overall data at the 0.01 significance level (see Appendix \ref{sec:mcnemar_test_results} for more details).

PCoT significantly improves disinformation detection across various scenarios, including news articles, social media posts, and novel post-cutoff datasets. It achieves the most substantial improvement in articles, with a 18\% increase. 
% highlighting its effectiveness in detecting disinformation in longer texts. 
Additionally, PCoT shows a 16\% improvement for post-cutoff datasets, leading to our next key finding:
\begin{tcolorbox}[
  colback=green!3!white, 
  colframe=green!40!black, 
  title=\textbf{\textcolor{white}{\small Finding}}, 
  boxsep=1pt, % Reduces padding inside the box
  left=1pt, % Reduces left margin
  right=1pt, % Reduces right margin
  top=1pt, % Reduces top padding
  bottom=1pt % Reduces bottom padding
]
\small Infusing persuasion knowledge into prompts improves generative LLMs' disinformation detection, especially for long texts and data not seen during pretraining.
\end{tcolorbox}
\noindent Better performance on unseen data confirms superior effectiveness on longer articles, as these datasets consist exclusively of such texts. We attribute PCoT's improved effectiveness on articles to the greater prevalence of persuasive strategies in longer texts, which complicate disinformation detection even for humans \cite{peng2024navigating}, underscoring the need for persuasion knowledge. Furthermore, PCoT deliver the largest average improvement, about 18\%, for the smallest model.
% , Llama 3.1 8B.

\begin{figure}[ht]  % Use a figure environment to control the placement
    \centering
    \includegraphics[width=\columnwidth]{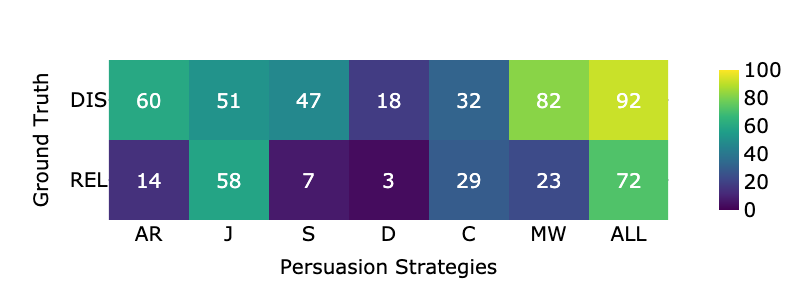}  % Replace with your image file name
    \caption{Average percentage of persuasion strategies predicted across 5 models for disinformation (\textit{DIS}) and reliable information (\textit{REL}). \textit{ALL} represents the percentage of instances with at least one detected persuasion strategy. Other abbreviations are explained in Figure \ref{fig:persuasion_strategies_tax}.}
    \label{fig:heatmap_true}
\end{figure}

\paragraph{Impact of Persuasion}
As Figure \ref{fig:heatmap_true} shows, at least one persuasion strategy was found in 92\% of disinformation and in 72\% of credible texts. These results suggest that persuasion is more commonly used in disinformation than in credible information, though a significant proportion of credible content also contains persuasion. The strongest correlation is observed between disinformation and the prediction of four specific strategies: \textit{Attack on reputation}, \textit{Simplification}, \textit{Distraction}, and \textit{Manipulative wording}. In contrast, the remaining two strategies, namely \textit{Justification} and \textit{Call}, occur with similar frequencies in both disinformation and credible information. More analysis confirming these findings can be found in Appendix \ref{sec:corelation}. The comparable presence of \textit{Call} and \textit{Justification} in both disinformation and credible content may be explained by the broad applicability of the persuasion techniques they encompass. For instance, \textit{Call} techniques like \textit{Slogans} such as "\textit{Make America Great Again!}" are highly persuasive but not inherently misleading, making them familiar across various types of content. Similarly, \textit{Conversation Killers} like "\textit{That’s just your opinion}" appear in discussions to shut down debate rather than mislead. Likewise, \textit{Justification} includes techniques often found in credible information. For instance, \textit{Appeal to Authority} is a standard persuasion technique in legitimate discourse, where expert opinions are cited to support claims. Similarly, \textit{Appeal to Popularity}, justifying an argument based on widespread acceptance can be found in factual contexts.

\begin{table}[ht]
\scriptsize
\centering
\renewcommand{\arraystretch}{1.1}
\setlength{\tabcolsep}{5pt} % Reduce horizontal spacing between columns
\begin{tabular}{lcccc}
\hline
\textbf{Model} & \multicolumn{2}{c}{\textbf{Persuasion}} & \multicolumn{2}{c}{\textbf{No Persuasion}} \\ \cline{2-5}
               & \textbf{PCoT}    & \textbf{Base}    & \textbf{PCoT}      & \textbf{Base}     \\ \hline
\multirow{2}{*}{GPT 4o Mini }   & \percnew{0.824}{0.872}         & 0.824         & \percnew{0.305}{0.342}           & 0.305          \\ 
  & \scriptsize $\pm$ 0,006 & \scriptsize $\pm$ 0,008 & \scriptsize $\pm$ 0,025 & \scriptsize $\pm$ 0,009  \\\hline
  
\multirow{2}{*}{Gemini 1.5 Flash} & \percnew{0.738}{0.844}       & 0.738         & \percnew{0.430}{0.444 }          & 0.430          \\ 
  & \scriptsize $\pm$ 0,014 & \scriptsize $\pm$ 0,036 & \scriptsize $\pm$ 0,013 & \scriptsize $\pm$ 0,007 \\\hline
\multirow{2}{*}{Claude 3 Haiku} & \percnew{0.756}{0.831}         & 0.756         & \percnew{0.295}{0.177}          & 0.295          \\ 
  & \scriptsize $\pm$ 0,014 & \scriptsize $\pm$ 0,101 & \scriptsize $\pm$ 0,043 & \scriptsize $\pm$ 0,084 \\\hline
\multirow{2}{*}{Llama 3.3 70B}  & \percnew{0.781}{0.871}         & 0.781         & \percnew{0.343}{0.409}           & 0.343          \\ 
  & \scriptsize $\pm$ 0,007 & \scriptsize $\pm$ 0,007 & \scriptsize $\pm$ 0,010 & \scriptsize $\pm$ 0,006  \\\hline
\multirow{2}{*}{Llama 3.1 8B}   & \percnew{0.679}{0.812}         & 0.679         & \percnew{0.494}{0.536}           & 0.494          \\ 
  & \scriptsize $\pm$ 0,008 & \scriptsize $\pm$ 0,050 & \scriptsize $\pm$ 0,014 & \scriptsize $\pm$ 0,059  \\\hline
  \textbf{Average}   & \percnew{0.753}{0.847}         & 0.753         & \percnew{0.368}{0.392}           & 0.368          \\ 
\end{tabular}
\caption{Results comparison across two subsets: \textit{Persuasion}, containing texts with at least one predicted persuasion strategy, and \textit{No Persuasion} texts with no predicted persuasion. The table reports the average F\textsubscript{1} score and standard deviation for each model across three different prompting methods.
}
\label{tab:predicted_persuasion_results}
\end{table}

In addition, we evaluated how the PCoT method enhances disinformation detection across two subsets of used datasets: one where at least one persuasion strategy was predicted and another where none was detected. As shown in Table \ref{tab:predicted_persuasion_results}, PCoT improves detection by an average of 12\% in the persuasion-present subset and about 7\% in the persuasion-absent subset (detailed results for individual prompting methods are provided in Appendix \ref{sec:persuasiove_non_persuasive_analysis}). 
Our findings highlight that: 
\begin{tcolorbox}[
  colback=green!3!white, 
  colframe=green!40!black, 
  title=\textbf{\textcolor{white}{\small Finding}}, 
  boxsep=1pt, % Reduces padding inside the box
  left=1pt, % Reduces left margin
  right=1pt, % Reduces right margin
  top=1pt, % Reduces top padding
  bottom=1pt % Reduces bottom padding
]
\small Detecting disinformation is particularly challenging in texts where no persuasion strategy has been predicted.
\end{tcolorbox}
\noindent Persuasive strategies may introduce emotionally charged language, making deception more apparent when these strategies are analyzed carefully. In contrast, when persuasion is absent, false statements alone may evade detection \cite{sosnowski2024eu}. In this scenario, fact-checking techniques become more crucial, and semantic analysis of the language alone may be insufficient.

\section{Further Evaluation and Ablation Study}

%In this section, we 
We present additional experiments: a comparison with other prompting methods (section~\ref{sec:additional_prompting}), a comparison of PCoT against cutting-edge reasoning models (section~\ref{sec:additional_reasoning}), 
%and an evaluation of explicit persuasion knowledge in zero-shot disinformation detection (section~\ref{sec:additional_ablation}).
and an ablation study to assess the impact of the definitions of the persuasion strategies to the overall performance (section~\ref{sec:additional_ablation}).

\subsection{Prompting Methods Comparison\label{sec:additional_prompting}}
We compare PCoT with other recent prompting methods, including CoT \cite{wei2022chain} in a zero-shot version (Z-CoT) \cite{lucas2023fighting}, Chain-of-Verification (CoVe) \cite{dhuliawala2024chain} and Rephrase and Respond (RaR) \cite{deng2023rephrase}. As shown in Table~\ref{tab:prompting_comparison}, PCoT consistently outperforms these methods.

\begin{table}[h]
%\scriptsize
\footnotesize
    \centering
    \begin{tabular}{lcccc}
        \toprule
        \textbf{Model} & \textbf{Z-CoT} & \textbf{RaR} & \textbf{CoVe} & \textbf{PCoT} \\
        \midrule
        GPT 4o Mini       & 0.765 & 0.698 & 0.790 & \textbf{0.846} \\
        Gemini 1.5 Flash  & 0.689 & 0.573 & 0.736 & \textbf{0.808} \\
        Claude 3 Haiku    & 0.588 & 0.768 & 0.441 & \textbf{0.774} \\
        Llama 3.3 70B     & 0.722 & 0.657 & 0.835 & \textbf{0.843} \\
        Llama 3.1 8B & 0.660 & 0.566 & 0.764 & \textbf{0.791} \\
        \bottomrule
    \end{tabular}
    \caption{Overall F\textsubscript{1} scores of different prompting methods on five datasets.}
    \label{tab:prompting_comparison}
\end{table}

\subsection{Evaluation Against Reasoning Models\label{sec:additional_reasoning}}

To further evaluate our approach, we compared PCoT-enhanced models to OpenAI’s advanced reasoning models, \textit{o1-mini} and \textit{o3-mini}. Specifically, we selected the best-performing (\textit{GPT-4o Mini}) and worst-performing (\textit{LLaMA 3.1 8B}) models from our zero-shot disinformation detection experiments using PCoT (see Table~\ref{tab:f1_overall_scores}) and compared them against the reasoning models.

As shown in Table~\ref{tab:reasoning_comparison}, even the weakest model, when used with PCoT, outperforms both \textit{o1-mini} and \textit{o3-mini} in zero-shot disinformation detection. This highlights PCoT’s ability to boost reasoning performance, even in smaller models.

\begin{table}[h]
    \centering
    %\scriptsize
    \footnotesize
    \begin{tabular}{lccccc}
        \toprule
        \textbf{Model} & \textbf{Overall} \\
        \midrule
        GPT 4o Mini + PCoT       & \textbf{0.846} \\
        Llama 3.1 8B + PCoT      & 0.791 \\
        o3-mini                  & 0.770  \\
        o1-mini                  & 0.634   \\
        \bottomrule
    \end{tabular}
    \caption{Overall F\textsubscript{1} scores for PCoT-enhanced models vs. OpenAI reasoning models on five datasets.}
    \label{tab:reasoning_comparison}
\end{table}

\subsection{PCoT Base Version and Ablation Results\label{sec:additional_ablation}}

To better understand the contribution of explicit persuasion knowledge in PCoT, we conducted an ablation study using a simplified base version, which provides in the prompt a general definition of persuasion avoiding to mention persuasion strategies.

Remarkably, even without detailed knowledge, this simplified version yields notable performance gains over baseline prompting methods across five datasets (see Table~\ref{tab:pcot_base_version}). Although the original variant of PCoT remains stronger, these findings underscore the role of persuasion-augmented reasoning in zero-shot disinformation detection.

%This variant removes the strategy-specific knowledge augmentation from \citet{piskorski2023multilingual} in the persuasion detection step. Instead, the model identifies the presence of persuasion and explains its role without referencing specific persuasion strategies. 

\begin{table}[ht]
\scriptsize
% \footnotesize
\centering
\renewcommand{\arraystretch}{1.1}
\begin{tabular}{lcc}
\hline
    \textbf{Model}           & \textbf{PCoT BV}    & \textbf{Base}       \\ \hline
GPT 4o Mini   & \percnew{0.765}{0.814} \scriptsize $\pm$ 0,007       & 0.765  \scriptsize $\pm$ 0,007 \\ 
Gemini 1.5 Flash & \percnew{0.705}{0.790}  \scriptsize $\pm$ 0,014    & 0.705   \scriptsize $\pm$ 0,034 \\ 
Claude 3 Haiku & \percnew{0.693}{0.736}   \scriptsize $\pm$ 0,013     & 0.693  \scriptsize $\pm$ 0,097 \\
Llama 3.3 70B  & \percnew{0.731}{0.831} \scriptsize $\pm$ 0,007      & 0.731 \scriptsize $\pm$ 0,009 \\ 
Llama 3.1 8B  & \percnew{0.661}{0.785}   \scriptsize $\pm$ 0,011     &  0.661 \scriptsize $\pm$ 0,035  \\ 
\hline
\end{tabular}
\caption{Comparison of average F\textsubscript{1} scores and standard deviations between Base prompts and PCoT without persuasion strategy augmentation. Results are shown for VaN, Z-CoT, and DeF-SpeC (as \textit{Base}), and their adaptations for PCoT's base version (\textit{PCoT BV}).
}
\label{tab:pcot_base_version}
\end{table}

\section{Related Works}

 \textbf{Disinformation Detection}. Disinformation detection has become a focal research focus due to its increasing impact on digital communication and societal trust \cite{flew2019digital, olan2024fake, iosifidis2020digital, martens2018digital}. Traditional approaches have used machine learning and deep learning to analyze lexical, semantic, and engagement-based features \cite{aslam2021fake, ali2022deep, nguyen2020fang}. Given the high stakes, explainability is crucial. Hybrid frameworks combining deep learning with feature-specific explanations enhance transparency, trust, and understanding in NLP applications \cite{hashmi2024advancing, reis2019explainable, cartwright2022detecting, shu2019defend}. Recent research has focused on detecting disinformation in human-annotated and LLM-generated data \cite{lucas2023fighting, chencan}. 

Limited annotated data has driven zero and few-shot learning, highlighting the adaptability of pre-trained transformers across tasks without domain-specific training \cite{sivarajkumar2023healthprompt, rizinski2023company, kumar2023zero, casola2023testing}. Researchers have shown that zero-shot detection with LLMs like GPT-4 can outperform supervised models like BERT in detecting disinformation \cite{pelrine2023towards, bang2023multitask, hassan2020political}.

\noindent\textbf{Datasets}. High-quality data is crucial for disinformation detection research \cite{d2021fake}, including datasets focused on COVID-19 disinformation \cite{bansal2021combining, cui2020coaid}, persuasion techniques \cite{da-san-martino-etal-2019-fine}, short statements \cite{wang2017liar} and fake news articles \cite{ahmed2018detecting, ahmed2017detection, shu2020fakenewsnet}. 
% A dataset with annotations of manipulation techniques was also created \cite{modzelewski2024mipd}. 
To the best of our knowledge, previous datasets are released without revealing the intermediate labels. In contrast, our MultiDis dataset includes annotations from each of the three annotation steps.

\noindent\textbf{Persuasion in Disinformation}. 
Different studies have shown that disinformation often uses persuasion and manipulation to mislead audiences \cite{modzelewski2024mipd, peng2023persuasive, chen2021persuasion, musi2022fallacies, ward2022identifying}. First attempts to use persuasion as intermediate labels in healthcare misinformation detection within a few-shot scenario have shown promising potential \cite{kamali2022using}. Nevertheless, no prior research has proposed a structured method that integrates persuasion knowledge and is applicable across diverse models and datasets to improve zero-shot disinformation detection.

\section{Conclusions}

In this study, we present \textbf{Persuasion-Augmented Chain of Thought (PCoT)}, a novel zero-shot approach that enhances disinformation detection by integrating persuasion knowledge into the LLM reasoning process. By leveraging persuasion knowledge and LLM-generated analysis, PCoT improves zero-shot classification, demonstrating the value of utilizing persuasion in disinformation detection.

Alongside PCoT, we present two novel disinformation datasets: MultiDis and EUDisinfo. EUDisinfo was collected using the database created by an EU initiative dedicated to identifying and analyzing pro-Kremlin disinformation. In contrast, MultiDis is a high-quality dataset created with debunking and fact-checking experts. %, following rigorous annotation. 
These datasets enabled a robust evaluation of PCoT on texts beyond the knowledge cutoff of tested LLMs (2024 onward). %from 2024 onward, beyond the knowledge cutoff of tested LLMs.

Our experiments using cutting-edge LLMs and five different datasets show that PCoT outperforms competitive methods, achieving an average 15\% improvement. PCoT enhances disinformation detection, particularly for longer texts, such as news articles, where it achieves a 18\% improvement while also providing a significant 8\% increase in accuracy for social media posts. 
We also identified four persuasion strategies that most correlate with disinformation:
%predicted persuasion strategies: 
\textit{Attack on reputation}, \textit{Simplification}, \textit{Distraction}, and \textit{Manipulative wording}. %, which correlate most with disinformation.

% Our findings highlight the critical role of persuasion-augmented reasoning in strengthening automated disinformation detection. It underscores the potential of persuasion-augmented AI to combat persuasive and deceptive online content.

\section*{Limitations}

\paragraph{Datasets and Annotation} Our annotation methodology for the MultiDis dataset categorized articles into one of eight thematic areas. Additionally, EUDisinfo focuses on disinformation with pro-Kremlin propaganda, while other datasets cover COVID-19 and political news disinformation. Despite this broad thematic coverage, we can not claim that the data fully represents all disinformation. Furthermore, our evaluation of the PCoT method was limited to English datasets. We leave multilingual analysis for future research.

\paragraph{Biases} Human annotation can be prone to subjectivity. To minimize bias, annotations in the \mbox{MultiDis} dataset were conducted in cooperation with experienced debunkers and fact-checkers. We developed comprehensive annotation guidelines and provided thorough training. Two independent annotators annotated each article, followed by a final third review by a senior supervisor. The supervisor consulted the initial annotator and/or a lead fact-checking expert to ensure accuracy and consistency when necessary. However, experiments involving externally annotated datasets inevitably inherited any biases in those sources.

\paragraph{Method} Our method heavily relies on the taxonomy and knowledge introduced by \citet{piskorski2023news}. Specifically, we incorporate a fixed set of high-level persuasive strategies in the first stage and integrate them into the prompt. 
Nevertheless, it is a taxonomy created by the Joint Research Centre, a scientific institution closely associated with the European Commission and used in many studies that confirm its high quality \cite{barron2024overview, dimitrov2024semeval, piskorski2023multilingual, szwoch2024limitations, leite2024cross}. Moreover, to the best of our knowledge, it is the only taxonomy extensively used for article annotation, which was crucial for our evaluation. A promising direction for future work is dynamically selecting persuasion techniques based on their relevance to disinformation detection. Nonetheless, we consider our current approach a crucial foundation for exploring dynamic selection, which we leave for future research.

\section*{Ethics}

\paragraph{Datasets and Annotation} The MultiDis and EUDisinfo datasets consist entirely of publicly available data with no copyright restrictions. They do not include any personally identifiable information and have been used exclusively for research purposes. The datasets will be released under the CC BY-NC-ND 4.0 license. Furthermore, our data collection protocol was reviewed and approved by an ethics board.

Crowdsourcing was not used at any stage of data collection or annotation. All annotators were employed by universities and fairly compensated. The annotation process remained entirely independent, free from political or commercial influence. Each team was overseen by two experienced supervisors and had direct access to a lead fact-checking expert for additional guidance.

\paragraph{Computational resources} Leveraging large language models often requires substantial computational resources, which can contribute to environmental concerns \cite{strubell2020energy}. However, our approach minimized computational demand as we relied on inference rather than training models from scratch. Most of our work conducted via usage of APIs, with no direct control over the computational resources involved. Additionally, we performed fine-tuning only on small BERT models. The computing infrastructure used for this research was acquired by the university specifically for research and educational purposes.

\section*{Acknowledgements}

This research was supported by the projects: Infotester4Education (full title: Development and implementation of AI education methods and digital tools supporting tackling disinformation, number:  2023-2-PL01-KA220-HED-000180856) within the framework of Cooperation Partnership for Higher Education, ERASMUS+, and EUonAIR project (number 101177370, ERASMUS-EDU-2024-EUR-UNIV-1), within the framework of the EUonAIR Centre of Excellence in Responsible AI in Education, co-funded by the European Commission.

Giovanni Da San Martino would like to thank the Qatar National Research Fund, part of Qatar Research Development and Innovation Council (QRDI), for funding this work  by grant NPRP14C0916-210015. 
He also would like to thank the European Union under the National Recovery and Resilience Plan (NRRP), Mission 4 Component 2 Investment 1.3 - Call for tender No. 341 of March 15, 2022 of Italian Ministry of University and Research – NextGenerationEU; Code PE00000014, Concession Decree No. 1556 of October 11, 2022 CUP D43C22003050001, Progetto "SEcurity and RIghts in the CyberSpace (SERICS) - Spoke 2 Misinformation and Fakes - DEcision supporT systEm foR cybeR intelligENCE (Deterrence) for also funding this work. 

% Bibliography entries for the entire Anthology, followed by custom entries
% \bibliography{anthology,custom}
% Custom bibliography entries only
% \bibliography{custom}

\appendix
\section{Dataset and Annotation Details}
\label{sec:methodology}
\subsection{Annotation Methodology and Guidelines}

Our methodology and annotation guidelines were designed to standardize the assessment of articles for disinformation content, aiming to reduce subjectivity and enable comprehensive analysis. Utilizing these annotation guidelines, we analyzed numerous articles to identify disinformation. The methodology was developed in cooperation with analysts (fact-checking and debunking experts) employed in the project based on their experience as experts, scientific knowledge available on the subject, and the experience of other institutions and organizations involved in research and detection of disinformation. The methodology improved throughout the project and subsequent testing to best reflect the disinformation environment. All authors of this methodology have at least three years of experience working for fact-checking or debunking organizations accredited by the International Fact-Checking Network. Moreover, our methodology and annotation guidelines draw on similar work on the annotation of disinformation, such as the guidelines presented by \citet{modzelewski2024mipd}.

\paragraph{Main Assumptions of the Methodology}
Creating a uniform methodology and guidelines aims to guarantee the quality of the assessments made by annotators and minimize their subjectivity. 

The analysis of articles is carried out mainly via the debunking technique, with the auxiliary use of the fact-checking technique. These terms for this methodology are defined in a manner analogous to the methodology developed for the NATO Strategic Communication Centre of Excellence \cite{pamment2021fact}. Fact-checking is the long-standing process of checking that all facts in a piece of writing, news article, or speech are correct. Debunking refers to exposing falseness or manipulating systematically and strategically (based on a chosen topic, classifications of selected techniques, narrative). 

\paragraph{Preparation of Articles for Evaluation} The first step is to select web portals from which articles on particular topics will be taken. Among them are both mainstream media and those presenting the alternative current. This is to ensure access to enough reliable as well as unreliable content. Each portal will be assigned to one of three categories, determining its credibility. This will be done by a team of experts by consensus. Assessing the credibility of a website requires an in-depth analysis of the content posted on it regularly, as well as checking it in reliable sources, including via the Media Bias/Fact Check search engine. The source's rating will not be visible to annotators. The analysis consists in selecting the category that best suits a given domain:
\begin{itemize}[nosep, leftmargin=*]
    \item {\textbf{Reliable}} — sources that are reliable/publishing reliable content on a specific topic, in particular traditional news portals.
    \item {\textbf{Unreliable}} — sources publishing unreliable content, typically disinformation, e.g., all domains financed by the Kremlin, sites containing conspiracy theories, etc.
    \item {\textbf{Mixed/Biased}} — partially or potentially biased websites that may present false information on specific issues, e.g., typically political websites, and blog collections.
\end{itemize}

\paragraph{Thematic Category}

Before the analysis begins, articles will be assigned to eight topics. This will be done manually with the help of keywords through searches on selected web portals. Thematic categories were pre-defined. The selection of topics was based on EU DisinfoLab's cross-cutting report on disinformation in Europe\cite{kn:Sessa}. It is based on expert studies from 20 countries. 

\begin{itemize}[nosep, leftmargin=*]
    \item Anti-Europeanism and anti-Atlanticism (anti-EU, anti-NATO)
    \item Anti-migration and xenophobia
    \item Climate change and the energy crisis 
    \item Health (including COVID-19 and vaccines)
    \item Institutional and media distrust (public institutions)
    \item Gender-based disinformation
    \item Ukraine war and refugees 
    \item Disinformation about LGBTQIA+
\end{itemize}

\paragraph{Content Analysis} 

The next step requires analyzing the entire article's content and recognizing whether the information is accurate or disinformative. If the article provides only factual information, it is marked as “credible information.” Selecting this category ends the assessment of the article. When information in the article is unreliable and misleads the recipients, content is considered disinformative. The unintentional dissemination of false information is known as misinformation. However, even unintentional dissemination of false information without the goal of manipulating recipients can fuel disinformation. Disinformation is particularly difficult to detect as the author’s intention is usually unspecified, and in most cases, it can only be presumed. Therefore, for this study, we assume that any form of false or manipulative information is considered disinformation.

For these guidelines, the definition of disinformation provided by the European Commission High-Level Group of Experts on False News and Disinformation on the Internet (HELG) will be used, as it covers all four aspects and does not exclude potentially harmful content presented in the form of political advertising or satire, as presented in the EU Code of Practice. The definition is as follows \cite{de2018multi}: 
\begin{quote}
    \itshape % Italicize the text
    `` All forms of false, inaccurate, or misleading information designed, presented, and promoted to intentionally cause public harm or for profit.''
    \hfill  % Add the citation on the right side
\end{quote}

However, a necessary supplement to this definition is taking into account the European Union Code of Practice on Disinformation, according to which disinformation is defined as: "verifiable false or misleading information which, cumulatively, (a) is created, presented and disseminated for economic gain or to intentionally deceive the public; and (b) may cause public harm, intended as threats to democratic political and policymaking processes as well as public goods such as the protection of EU citizens' health, the environment, or security". \cite{eu2022strengthdiscode}. The detected information must be verifiable, which means that it can be proved untrue, and, therefore, it cannot be, for example, a yet unproven theory or opinion, as long as it is not intended to mislead the recipients. In summary, disinformation is intentionally misleading by providing misleading or false information \cite{eu2020communication}. Unlike disinformation, misinformation is \textit{misleading information shared by people who do not recognize it as such} \cite{de2018multi}. However, as noted earlier, misinformation and disinformation are treated as a single category under "disinformation."

When a given content is not verifiable (reliable/disinformative/misinformative), it is marked as the "Hard to say" category. Indicating this category ends the assessment the same as "Inconsistent with the topic". Below, we present the main categories:

\begin{itemize}[nosep, leftmargin=*]
    \item Credible information
    \item Disinformation 
    \item Hard to say
    \item Inconsistent with the topic 
\end{itemize}

\section{Dataset Basic Statistics}

This appendix provides additional details on the statistics of the two datasets created for this study. The second section presents statistics on the balance of the classes in the five test datasets. All datasets, including the test datasets used in our experiments, are available in our public repository\footnote{Repository with data, prompts and codebase: \url{https://github.com/ArkadiusDS/PCoT}}.

\subsection{MultiDis and EUDisinfo Statistics}
\label{sec:app_data_stats}

Table \ref{tab:articles_per_credibility} reports the number of articles within the three credibility categories in MultiDis dataset: \textit{Credible Information}, \textit{Disinformation}, and \textit{Hard-to-say}. Additionally, Table \ref{tab:articles_per_credibility} includes articles labeled as \textit{Inconsistent with the Topic}. Articles in this category were excluded from further analysis. Similar statistics for the EUDisinfo dataset are presented in Table \ref{tab:eudisinfo_articles}, which includes only two categories: \textit{Credible Information} and \textit{Disinformation}.

\begin{table}[ht]
    \centering
    \renewcommand{\arraystretch}{1.2} % Increase vertical spacing for readability
    \setlength{\tabcolsep}{5pt} % Reduce horizontal spacing between columns
    \small
    \begin{tabular}{l c c}
        \toprule
        \textbf{Category} & \textbf{\textit{\#DOC}} & \textbf{\textit{\#PERC}} \\
        \midrule
        Credible Information & 1256 & 65.3\% \\
        Disinformation & 630 & 32.8\% \\
        Hard-to-say & 18 & 0.95\% \\
        Inconsistent with the Topic & 18 & 0.95\% \\
        \bottomrule
    \end{tabular}
    \caption{Number of articles (\textit{\#DOC}) per each credibility evaluation category and their (\#PERC) percentage in MultiDis dataset.}
    \label{tab:articles_per_credibility}
\end{table}

\begin{table}[h!]
\centering
\small
\begin{tabular}{lcc}
\toprule
\textbf{Category} & \textbf{\#DOC} & \textbf{\#PERC} \\
\midrule
Credible Information & 241 & 67.1\% \\
Disinformation       & 118 & 32.9\% \\
\bottomrule
\end{tabular}
\caption{Number of articles (\#DOC) per main credibility evaluation category and their (\#PERC) percentage in the EUDisinfo test dataset.}
\label{tab:eudisinfo_articles}
\end{table}

\subsection{Test Datasets Statistics}
\label{sec:test_data_stats}

Table \ref{tab:dataset-distribution-transposed} presents the class distribution of \textit{Disinformation} and \textit{Credible Information} across the five test datasets. Table \ref{tab:category-distribution-transposed} shows the same distribution across different content categories and time-based splits, indicating that social media posts and prior cutoff texts contain a higher proportion of disinformation.

\begin{table}[h]
\centering
\scriptsize
\begin{tabular}{lcc}
\toprule
\textbf{Dataset} & \textbf{Disinformation} & \textbf{Credible Information} \\
\midrule
CoAID & 21\% & 79\% \\
ECTF & 41\% & 59\% \\
EUDisinfo & 33\% & 67\% \\
ISOT Fake News & 55\% & 45\% \\
MultiDis & 26\% & 74\% \\
\bottomrule
\end{tabular}
\caption{Class distribution across evaluation datasets. The proportions reflect the nature of each dataset and its composition regarding disinformation and credible content.}
\label{tab:dataset-distribution-transposed}
\end{table}

\vspace{1em}

\begin{table}[h]
\centering
\scriptsize
\begin{tabular}{lcc}
\toprule
\textbf{Category} & \textbf{Disinformation} & \textbf{Credible Information} \\
\midrule
All Texts & 35\% & 65\% \\
Articles & 33\% & 67\% \\
Social Media Posts & 41\% & 59\% \\
Prior Cutoff & 39\% & 61\% \\
Post Cutoff & 29\% & 71\% \\
\bottomrule
\end{tabular}
\caption{Class distribution by text type and time period. Social media and pre-cutoff texts show a higher share of disinformation compared to articles and post-cutoff samples.}
\label{tab:category-distribution-transposed}
\end{table}

\section{Persuasion Strategies and Techniques}
\label{sec:persuasion_strategies_techniques}
The six general persuasion strategies in our study are linked to specific persuasion techniques, as identified by \citet{piskorski2023multilingual, piskorski2023news}. Definitions of these techniques are provided in the final prompt created for the first stage of PCoT method.
\subsection{Attack on Reputation}
\begin{itemize}[nosep, leftmargin=*]
    \item \textbf{Name Calling or Labelling}: a form of argument in which loaded labels are directed at an individual, group, object or activity, typically in an insulting or demeaning way, but also using labels the target audience finds desirable.
    \item \textbf{Guilt by Association}: attacking the opponent or an activity by associating it with another group, activity or concept that has sharp negative connotations for the target audience.
    \item \textbf{Casting Doubt}: questioning the character or personal attributes of someone or something in order to question their general credibility or quality.
    \item \textbf{Appeal to Hypocrisy}: the target of the technique is attacked on its reputation by charging them with hypocrisy/inconsistency.
    \item \textbf{Questioning the Reputation}: the target is attacked by making strong negative claims about it, focusing specially on undermining its character and moral stature rather than relying on an argument about the topic.
\end{itemize}

\subsection{Justification}
\begin{itemize}[nosep, leftmargin=*]
    \item \textbf{Flag Waving}: justifying an idea by exhaling the pride of a group or highlighting the benefits for that specific group.
    \item \textbf{Appeal to Authority}: a weight is given to an argument, an idea or information by simply stating that a particular entity considered as an authority is the source of the information.
    \item \textbf{Appeal to Popularity}: a weight is given to an argument or idea by justifying it on the basis that allegedly "everybody" (or the large majority) agrees with it or "nobody" disagrees with it.
    \item \textbf{Appeal to Values}: a weight is given to an idea by linking it to values seen by the target audience as positive.
    \item \textbf{Appeal to Fear, Prejudice}: promotes or rejects an idea through the repulsion or fear of the audience towards this idea.
\end{itemize}

\subsection{Distraction}
\begin{itemize}[nosep, leftmargin=*]
    \item \textbf{Strawman}: consists in making an impression of refuting an argument of the opponent’s proposition, whereas the real subject of the argument was not addressed or refuted, but instead replaced with a false one.
    \item \textbf{Red Herring}: consists in diverting the attention of the audience from the main topic being discussed, by introducing another topic, which is irrelevant.
    \item \textbf{Whataboutism}: a technique that attempts to discredit an opponent’s position by charging them with hypocrisy without directly disproving their argument.
\end{itemize}

\subsection{Simplification}
\begin{itemize}[nosep, leftmargin=*]
    \item \textbf{Causal Oversimplification}: assuming a single cause or reason when there are actually multiple causes for an issue.
    \item \textbf{False Dilemma or No Choice}: a logical fallacy that presents only two options or sides when there are many options or sides. In extreme, the author tells the audience exactly what actions to take, eliminating any other possible choices.
    \item \textbf{Consequential Oversimplification}: is an assertion one is making of some "first" event/action leading to a domino-like chain of events that have some significant negative (positive) effects and consequences that appear to be ludicrous or unwarranted or with each step in the chain more and more improbable.
\end{itemize}

\subsection{Call}
\begin{itemize}[nosep, leftmargin=*]
    \item \textbf{Slogans}: a brief and striking phrase, often acting like emotional appeals, that may include labeling and stereotyping.
    \item \textbf{Conversation Killer}: words or phrases that discourage critical thought and meaningful discussion about a given topic.
    \item \textbf{Appeal to Time}: the argument is centred around the idea that time has come for a particular action.
\end{itemize}

\subsection{Manipulative Wording}
\begin{itemize}[nosep, leftmargin=*]
    \item \textbf{Loaded Language}: use of specific words and phrases with strong emotional implications (either positive or negative) to influence and convince the audience that an argument is valid.
    \item \textbf{Obfuscation, Intentional Vagueness, Confusion}: use of words that are deliberately not clear, vague or ambiguous so that the audience may have its own interpretations.
    \item \textbf{Exaggeration or Minimisation}: consists of either representing something in an excessive manner or making something seem less important or smaller than it really is.
    \item \textbf{Repetition}: the speaker uses the same phrase repeatedly with the hopes that the repetition will lead to persuade the audience.
\end{itemize}

\section{LLMs Explored in Experiments}
\label{sec:models}

\begin{table*}[ht!]
\scriptsize
\centering
\begin{tabular}{lllll}
\toprule
 \textbf{API Model Name} & \textbf{Knowledge Cutoff Date} & \textbf{Access Details} & \textbf{License} & \textbf{Model Size}\\\midrule
 \texttt{gpt-4o-mini}&October 2023 & OpenAI API 02.2025 & Commercial & Not Disclosed\\
 \texttt{gemini-1.5-flash}& November 2023 & Google API 02.2025 & Commercial & Not Disclosed\\ 
 \texttt{claude-3-haiku-20240307} & August 2023 & Anthropic API 02.2025 & Commercial& Not Disclosed\\ 
 \texttt{meta-llama/Llama-3.3-70B-Instruct-Turbo}&  December 2023 & DeepInfra API 02.2025 & Meta Llama 3 Community& 70B \\ 
 \texttt{meta-llama/Meta-Llama-3.1-8B-Instruct}& December 2023 & DeepInfra API 02.2025 & Meta Llama 3 Community& 8B \\ 
\bottomrule
\end{tabular}
\caption{Large Language Models used in our experiments.}
\label{tab:llm-models}
\end{table*}

In our experiments, we used five different cutting-edge LLMs: \textit{GPT-4o-mini}, \textit{Meta-Llama 3.1} (8B-Instruct), \textit{Claude 3 Haiku}, \textit{Llama 3.3} (70B-Instruct-Turbo), and \textit{Gemini 1.5 Flash}. We aimed to include widely recognized, state-of-the-art models from the largest available while ensuring they remain affordable. We also selected two open-weight models to demonstrate that our method can be applied without access to closed models through APIs. Additionally, we chose the smaller Llama 3.1 with 8B parameters to ensure that our method could be applied to models that do not require costly infrastructure.
 
 Table \ref{tab:llm-models} lists the Large Language Models used in our experiments, detailing their knowledge cutoff dates, access methods, licenses, and sizes. The knowledge cutoff dates confirm that our datasets, \textit{MultiDis} and \textit{EUDisinfo}, which contain articles from 2024 onward, were not part of the models' pretraining.

\section{Persuasion Detection Step Evaluation and Impact on PCoT}
\label{sec:persuasion_det_step_exp_and_impact_on_pcot}
We conducted a series of extensive experiments to optimize the first stage of our PCoT method. 
Our experiments extensively relied on the taxonomy developed by \citet{piskorski2023news, piskorski2023multilingual}, particularly for crafting prompts to detect persuasive strategies. The taxonomy was developed by researchers at the Joint Research Centre (JRC)\footnote{\href{https://commission.europa.eu/about/departments-and-executive-agencies/joint-research-centre_en}{JRC} is a scientific institution closely associated with the European Commission. The center provides independent, evidence-based insights and research to support the EU for societal benefit.}. This choice was driven by the fact that the taxonomy and the annotated data are publicly available. These datasets were used in the International Workshop on Semantic Evaluation, focusing on persuasion detection \cite{piskorski2023semeval, dimitrov2024semeval}. By leveraging this resource, we could evaluate our taxonomy using ground truth data. Additionally, to our knowledge, this is the only high-quality taxonomy applied to longer-form news articles. News articles are a key component of our extensive experiments.

Below, we provide a description of the five key approaches we tested with shortcut in brackets:
\begin{enumerate}[nosep, leftmargin=*] 
    \item \textbf{Multitask [MT]} -  In this approach, we used a single prompt that included the names and definitions of persuasion strategies, as outlined by \citet{piskorski2023multilingual} and showed in Figure \ref{fig:persuasion_strategies_tax}. This zero-shot prompting method guided the LLM in classifying persuasion strategies across multiple labels and categories. Furthermore, we instructed the model to provide explanations for each classification decision.
    \item \textbf{Detailed Multitask [DMT]} - In this approach, we used a single, comprehensive prompt that provided additional context for each persuasion strategy to improve the performance of the classification tasks. Along with the definitions of the persuasion strategies, we included various techniques related to each strategy, with their definitions outlined by \citet{piskorski2023multilingual}. Specifically, this method incorporated the names and definitions of strategies listed in Figure \ref{fig:persuasion_strategies_tax} and the names and definitions of techniques from Section \ref{sec:persuasion_strategies_techniques}. Furthermore, we requested that the model explain each decision made in classifying persuasion strategies.
    \item \textbf{One Task at a Time [TAT]} - In this approach, we used a separate prompt for each persuasion strategy, treating each as a binary classification task. This approach resulted in six distinct prompts, each focusing on a specific persuasion strategy from \citet{piskorski2023multilingual}. Each prompt included only the name and definition of a single strategy, as listed in Figure \ref{fig:persuasion_strategies_tax}. Additionally, we asked the model to explain each classification decision related to the corresponding persuasion strategy.
    \item \textbf{One Detailed Task at a Time [DTAT]} - This approach is similar to the \textit{One Task at a Time} method but with more detailed information to aid in the binary classification of each persuasion strategy. For each strategy, we used a separate prompt that not only included the name and definition of the strategy, as listed in Figure \ref{fig:persuasion_strategies_tax}, but also provided the names and definitions of the persuasion techniques associated with that strategy, as outlined in Section \ref{sec:persuasion_strategies_techniques}. As with the other approaches, we followed the taxonomy from \citet{piskorski2023multilingual} to structure the prompts.
    \item \textbf{One Task at a Time with Broad Knowledge [TATB]} - This approach is similar to the One Task at a Time method but with a broader scope. Instead of providing knowledge about a single persuasion strategy per prompt, we used six distinct prompts, each containing knowledge about all the persuasion strategies. However, the LLM was still tasked with detecting and analyzing only one specific strategy within each prompt, treating it as a binary classification task.
\end{enumerate}

\begin{table}[ht]
    \centering
    \scriptsize % Make the font size smaller
    \begin{tabular}{l c}
        \toprule
        Prompting Method & F\textsubscript{1} Score \\
        \midrule
        PCoT DMT & 0.815 {$\uparrow$} 1.6 p.p. \\
        PCoT No Exp & 0.799 {$\uparrow$} 3.4 p.p. \\
        PCoT Single Step & 0.765 {$\uparrow$} 5.4 p.p.\\
        Base & 0.711 \\
                \bottomrule
    \end{tabular}
    \caption{Impact of different prompting methods on final PCoT method performance}
    \label{tab:prompting_methods}
\end{table}

\begin{table}[ht]
\scriptsize % Make the font size smaller
\renewcommand{\arraystretch}{1.2} % Increase vertical spacing for readability
\setlength{\tabcolsep}{5pt} % Reduce horizontal spacing between columns
    \centering
    \begin{tabular}{llll}
    \toprule
    model & Explanation & F\textsubscript{1} Score \\
    \midrule
        \multirow[t]{10}{*}{GPT 4o mini} & Yes & 0.841 \\
     &   No & 0.830 \\
    \hline
    \multirow[t]{10}{*}{Gemini 1.5 Flash}  & Yes & 0.817 \\
     &   No & 0.798 \\
    \hline
    \multirow[t]{10}{*}{Claude 3 Haiku}  & Yes & 0.789 \\
     &   No & 0.771 \\
     \hline
    \multirow[t]{10}{*}{Llama 3.3 70B}  & Yes & 0.844 \\
     &   No & 0.842 \\
    \hline
    \multirow[t]{10}{*}{Llama 3.1 8B}  & Yes & 0.785 \\
     &   No & 0.756 \\
    \hline
    \multirow[t]{10}{*}{\textbf{Average}}  & Yes & 0.815  \\
     &   No & 0.799 \\
    \bottomrule
    \end{tabular}
    \caption{Results for PCoT with usage of explanation for each persuasion strategy and without explanation.}
    \label{tab:explanation}
\end{table}

\begin{table*}[ht]
\scriptsize % Make the font size smaller
\renewcommand{\arraystretch}{1.2} % Increase vertical spacing for readability
\setlength{\tabcolsep}{5pt} % Reduce horizontal spacing between columns
\centering
\begin{tabular}{|l|c|c|c|c|c|c|c|c|}
\hline
\textbf{Approach} & \textbf{Metric} & \textbf{Attack on Reputation} & \textbf{Justification} & \textbf{Simplification} & \textbf{Distraction} & \textbf{Call} & \textbf{Manipulative Wording} & \textbf{Average} \\ \hline

% \multirow{1}{*}
{MT} & F1 Micro & 0.6407 & 0.6616 & 0.6198 & 0.7537 & 0.6366 & 0.7813 & 0.6823 \\
 & & \scriptsize $\pm$ 0.130 & \scriptsize $\pm$ 0.031 & \scriptsize $\pm$ 0.022 & \scriptsize $\pm$ 0.090 & \scriptsize $\pm$ 0.046 & \scriptsize $\pm$ 0.093 & \scriptsize $\pm$ 0.069 \\ \cline{2-9}
 % & F1 Macro & 0.6104 & 0.5775 & 0.5755 & 0.5257 & 0.5921 & 0.6409 & 0.5870 \\
 % & & \scriptsize $\pm$ 0.107 & \scriptsize $\pm$ 0.057 & \scriptsize $\pm$ 0.041 & \scriptsize $\pm$ 0.016 & \scriptsize $\pm$ 0.025 & \scriptsize $\pm$ 0.073 & \scriptsize $\pm$ 0.053 \\ \cline{2-9}
 % & F1 Weighted & 0.6500 & 0.6349 & 0.6040 & 0.7561 & 0.6336 & 0.8217 &0.6834 \\
 % & & \scriptsize $\pm$ 0.135 & \scriptsize $\pm$ 0.036 & \scriptsize $\pm$ 0.027 & \scriptsize $\pm$ 0.053 & \scriptsize $\pm$ 0.040 & \scriptsize $\pm$ 0.070 & \scriptsize $\pm$ 0.060 \\ \hline

% \multirow{1}{*}
{DMT} & F1 Micro & 0.7368 & 0.6710 & 0.6290 & 0.7082 & 0.6326 & 0.8440 & 0.7036 \\
 & & \scriptsize $\pm$ 0.103 & \scriptsize $\pm$ 0.044 & \scriptsize $\pm$ 0.028 & \scriptsize $\pm$ 0.104 & \scriptsize $\pm$ 0.046 & \scriptsize $\pm$ 0.058 & \scriptsize $\pm$ 0.081 \\ \cline{2-9}
 % & F1 Macro & 0.6860 & 0.6022 & 0.6068 & 0.5464 & 0.5940 & 0.6916 & 0.6212 \\
 % & & \scriptsize $\pm$ 0.077 & \scriptsize $\pm$ 0.049 & \scriptsize $\pm$ 0.029 & \scriptsize $\pm$ 0.035& \scriptsize $\pm$ 0.027 & \scriptsize $\pm$ 0.053& \scriptsize $\pm$ 0.057 \\ \cline{2-9}
 % & F1 Weighted & 0.7446 & 0.6520 & 0.6246 & 0.7344 & 0.6310 & 0.8678 & 0.7091 \\
 % & & \scriptsize $\pm$ 0.094 & \scriptsize $\pm$ 0.038 & \scriptsize $\pm$ 0.027 & \scriptsize $\pm$ 0.075 & \scriptsize $\pm$ 0.041 & \scriptsize $\pm$ 0.041 & \scriptsize $\pm$ 0.093 \\ \hline

% \multirow{3}{*}
{TAT} & F1 Micro & 0.6522 & 0.6489 & 0.5940 & 0.5153 & 0.6541 & 0.7276 & 0.6320 \\
 & & \scriptsize $\pm$ 0.143 & \scriptsize $\pm$ 0.041 & \scriptsize $\pm$ 0.026 & \scriptsize $\pm$ 0.188 & \scriptsize $\pm$ 0.011 & \scriptsize $\pm$ 0.217 & \scriptsize $\pm$ 0.065 \\ \cline{2-9}
 % & F1 Macro & 0.6198 & 0.6022 & 0.5314 & 0.4144 & 0.6002 & 0.5944 & 0.5604 \\
 % & & \scriptsize $\pm$ 0.121 & \scriptsize $\pm$ 0.020 & \scriptsize $\pm$ 0.069 & \scriptsize $\pm$ 0.078 & \scriptsize $\pm$ 0.023 & \scriptsize $\pm$ 0.153 & \scriptsize $\pm$ 0.071 \\ \cline{2-9}
 % & F1 Weighted & 0.6582 & 0.6413 & 0.5554 & 0.5416 & 0.6488 & 0.7631 & 0.6347 \\
 % & & \scriptsize $\pm$ 0.153 & \scriptsize $\pm$ 0.024 & \scriptsize $\pm$ 0.042 & \scriptsize $\pm$ 0.173 & \scriptsize $\pm$ 0.012 & \scriptsize $\pm$ 0.185 & \scriptsize $\pm$ 0.073 \\ \hline

 % \multirow{3}{*}
 {DTAT} & F1 Micro & 0.6963 & 0.6896 & 0.5985 & 0.4810 & 0.6407 & 0.8045 & 0.6518 \\
 & & \scriptsize $\pm$ 0.086 & \scriptsize $\pm$ 0.013 & \scriptsize $\pm$ 0.028 & \scriptsize $\pm$ 0.147 & \scriptsize $\pm$ 0.023 & \scriptsize $\pm$ 0.100 & \scriptsize $\pm$ 0.109 \\ \cline{2-9}
 % & F1 Macro & 0.6593 & 0.6141 & 0.5738 & 0.4246 & 0.6016 & 0.6443 & 0.5863 \\
 % & & \scriptsize $\pm$ 0.066 & \scriptsize $\pm$ 0.033 & \scriptsize $\pm$ 0.032 & \scriptsize $\pm$ 0.085 & \scriptsize $\pm$ 0.017 & \scriptsize $\pm$ 0.069 & \scriptsize $\pm$ 0.085 \\ \cline{2-9}
 % & F1 Weighted & 0.7112 & 0.6681 & 0.5837 & 0.5212 & 0.6404 & 0.8354 & 0.6600 \\
 % & & \scriptsize $\pm$ 0.080 & \scriptsize $\pm$ 0.021 & \scriptsize $\pm$ 0.031 & \scriptsize $\pm$ 0.153 & \scriptsize $\pm$ 0.008 & \scriptsize $\pm$ 0.071 & \scriptsize $\pm$ 0.109 \\ \hline

 % \multirow{3}{*}
 {TATB} & F1 Micro & 0.6455 & 0.6437 & 0.5851 & 0.5269 & 0.6299 & 0.6858 & 0.6195 \\
 & & \scriptsize $\pm$ 0.133 & \scriptsize $\pm$ 0.045 & \scriptsize $\pm$ 0.047 & \scriptsize $\pm$ 0.248 & \scriptsize $\pm$ 0.058 & \scriptsize $\pm$ 0.225 & \scriptsize $\pm$ 0.056 \\ \cline{1-9}
 % & F1 Macro & 0.6105 & 0.5938 & 0.5008 & 0.3950 & 0.5376 & 0.5576 & 0.5326 \\
 % & & \scriptsize $\pm$ 0.112 & \scriptsize $\pm$ 0.032 & \scriptsize $\pm$ 0.081 & \scriptsize $\pm$ 0.132 & \scriptsize $\pm$ 0.040 & \scriptsize $\pm$ 0.151 & \scriptsize $\pm$ 0.078 \\ \cline{2-9}
 % & F1 Weighted & 0.6529& 0.6378 & 0.5315 & 0.5238& 0.5976 & 0.7270 & 0.6118 \\
 % & & \scriptsize $\pm$ 0.145 & \scriptsize $\pm$ 0.036 & \scriptsize $\pm$ 0.062 & \scriptsize $\pm$ 0.258 & \scriptsize $\pm$ 0.042 & \scriptsize $\pm$ 0.190 & \scriptsize $\pm$ 0.077 \\ \hline

\end{tabular}
\caption{The table presents F\textsubscript{1} scores for each persuasion strategy and approach. Standard deviations, calculated from the results across five different LLMs, are provided below their corresponding scores. Detailed explanations of each approach can be found in Section \ref{sec:persuasion_det_step_exp_and_impact_on_pcot}. The final approach used in the PCoT method is the best-performing \textit{DMT}.}
\label{tab:persuasion_results_approaches}
\end{table*}

 Table \ref{tab:persuasion_results_approaches} presents the average results for each of the described methods. It includes the overall average performance in detecting persuasion and the results for each persuasion strategy. The \textit{Detailed Multitask [DMT]} method outperformed the others in detecting persuasion. As a result, we selected \textit{DMT} for the final version of the first stage of our PCoT method.

The results in Table \ref{tab:prompting_methods} underscore the impact of different prompting strategies within the PCoT method for disinformation detection. The best-performing variant, \textit{PCoT DMT}, achieves an F\textsubscript{1} score of 0.815, surpassing the baseline by 10.4 percentage points. Excluding explanations of the persuasion strategy in the first stage (\textit{PCoT No Exp}) reduces the performance to 0.799, while adopting a single-stage approach (\textit{PCoT Single Step}) further reduces it to 0.765. These findings emphasize the critical role of a two-stage reasoning process and persuasion strategy analysis in enhancing disinformation detection.

Table \ref{tab:explanation} presents a comparative analysis of PCoT with and without persuasion strategy explanations across various models. The impact of explanations varies, with the most significant improvement observed in the smallest open-weight model, Llama 3.1 8B, while Llama 3.3 70B shows minimal change. We observe a consistent average improvement when using explanations. Since inference is conducted with a temperature of 0, making the results more stable and reproducible, this further reinforces the importance of explanations. Notably, the benefits are most pronounced for smaller models, underscoring the value of explanations in enhancing their disinformation detection performance.

\section{Prompts}
\label{sec:prompts}
In this section, we provide an overview of the prompts used in our study and present prompt templates for each step of the PCoT method. Given the large number and substantial length of the prompts, we do not include them in full in the paper. Instead, the complete set of prompts is available in our online repository.

\subsection{Baselines}
\label{sec:prompt_baseline}
\begin{figure*}[ht]  % Use a figure environment to control the placement
    \centering
    \includegraphics[width=1\textwidth]{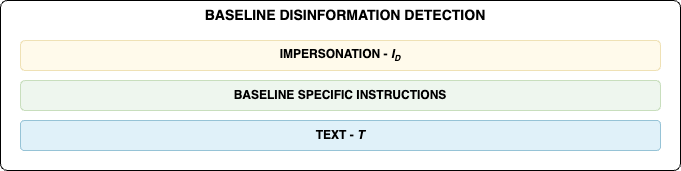}  % Replace with your image file name
    \caption{The prompt template for each baseline method in disinformation detection, namely, \textit{VaN}, \textit{Z-CoT}, and \textit{DeF-SpeC}. The component \(I_D\) establishes context while overriding alignment tuning. Each baseline method differs in the \textit{Baseline Specific Instructions} block. Generally, it provides method-specific guidelines defining the task and requests for structured output. Finally, the text \(T\) represents the content passed for disinformation evaluation.}
    \label{fig:base_disinformation_detection_prompt}
\end{figure*}

Figure \ref{fig:base_disinformation_detection_prompt} illustrates the baseline prompt template used for zero-shot disinformation detection, specifically for the \textit{VaN}, \textit{Z-CoT}, and \textit{DeF-SpeC} methods introduced by \citet{lucas2023fighting}. These methods were selected because \citet{lucas2023fighting} comprehensively evaluated various approaches using disinformation datasets, testing prompts on human-annotated and LLM-generated data. Since our study focuses exclusively on human-annotated data, we chose three of the best-performing methods on human-annotated data.

\subsection{Persuasion Detection Step}
\label{sec:prompt_persuasion_det}
\begin{figure*}[ht]  % Use a figure environment to control the placement
    \centering
    \includegraphics[width=1\textwidth]{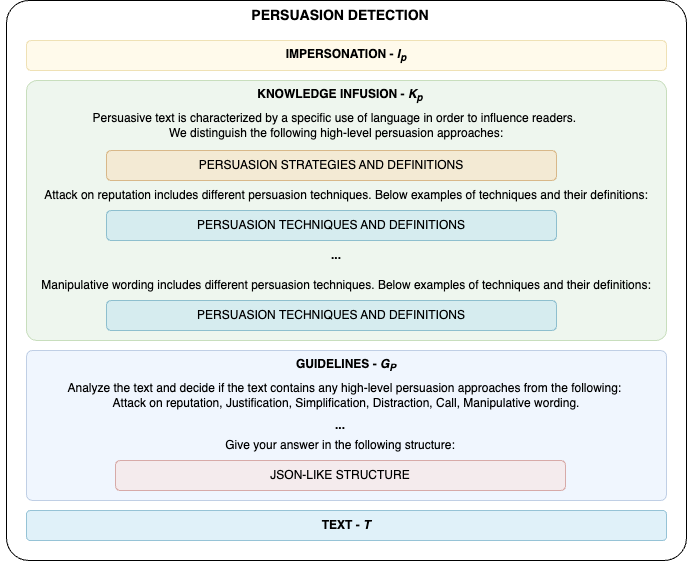}  % Replace with your image file name
\caption{The prompt template for first stage of PCoT method, namely for persuasion detection step. The component \( I_P \) establishes the context and overrides alignment tuning, while \( K_P \) encapsulates knowledge about a predefined set of high-level persuasion strategies, and guidelines \( G_P\) determine the task and specify the structure of the expected response. The \textit{Persuasion Strategies and Definitions} block includes names of persuasion strategies and definitions presented in Figure \ref{fig:persuasion_strategies_tax}, while \textit{Persuasion Techniques and Definitions} blocks includes names and definitions of techniques described in Appendix \ref{sec:persuasion_strategies_techniques}. Finally, the text \(T\) represents the content passed for persuasion analysis.}
    \label{fig:persuasion_detection_prompt}
\end{figure*}

Figure \ref{fig:persuasion_detection_prompt} presents the final template of the best-performing prompt used in the first stage of the PCoT method, designed specifically for detecting persuasion strategies and generating corresponding explanations. This prompt was meticulously crafted following a comprehensive evaluation of various approaches applied to data with ground truth labels for persuasion strategies. In particular, we tested multiple methods on the dataset from the International Workshop on Semantic Evaluation 2023 (SemEval 2023) shared task on persuasion \cite{piskorski2023semeval}. The final prompt incorporates the names and definitions of persuasive strategies and the associated techniques outlined in \citet{piskorski2023multilingual, piskorski2023news}. Figure \ref{fig:persuasion_detection_prompt} offers a detailed view of the prompt used in our study. Additionally, we make the final prompts publicly available.

\subsection{Disinformation Detection Step}
\label{sec:prompt_pcot_final}
\begin{figure*}[ht]  % Use a figure environment to control the placement
    \centering
    \includegraphics[width=1\textwidth]{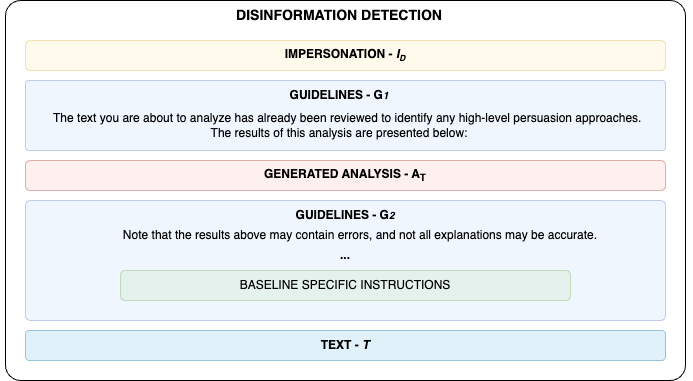}  % Replace with your image file name
\caption{The prompt template for second final stage of PCoT method, namely for disinformation detection step. The component \( I_D \) establishes the context and overrides alignment tuning, while guidelines \( G_D = \{G_1,G_2\} \) determine the task and specify the structure of the expected response. Next component is the generated analysis \( A_T \) from the output of first stage of PCoT and finally, the text \(T\) represents the content passed for disinformation evaluation. The \textit{Baseline Specific Instructions} block is a part of guidelines and includes different instructions depending on which baseline method was adapted to PCoT method, namely it can be instruction from \textit{VaN}, \textit{Z-CoT}, or \textit{DeF-SpeC}}
    \label{fig:disinformation_detection_prompt}
\end{figure*}

Figure \ref{fig:disinformation_detection_prompt} illustrates the final prompt template used in the second stage of the PCoT method, which focuses on disinformation detection. This prompt incorporates the persuasion analysis generated in the first stage of PCoT. For each test set, we experimented with three different disinformation prompts. We adjusted three methods \textit{VaN}, \textit{Z-CoT}, and \textit{DeF-SpeC} \cite{lucas2023fighting} to our PCoT method. This approach enabled us to compare the performance of the adapted methods against the baselines, where we applied the original methods from \citet{lucas2023fighting}.

\section{Detailed Analysis}

\subsection{McNemar's Test for PCoT Performance}
\label{sec:mcnemar_test_results}
To evaluate the statistical significance of PCoT, we conducted McNemar's test comparing each prompting method to its PCoT-adjusted counterpart across various language models. The results, presented in Table~\ref{tab:mc_nemars_test}, show that PCoT consistently improves performance at the 0.01 significance level across all models and methods in overall evaluation. However, certain cases, such as experiments on posts for Llama 3.1 8B and experiments on articles for Claude 3 Haiku in DeF-Spec, exhibit non-significant differences.

\subsection{Persuasion Strategy Correlations with Disinformation Detection}
\label{sec:corelation}
The results presented in Tables \ref{tab:matthews_correlation} and \ref{tab:matthews_correlation_detailed} provide key insights into the relationship between persuasion strategies and disinformation detection across different models and prompting methods. Table \ref{tab:matthews_correlation} presents the Matthews correlation coefficient (MCC) between various persuasion strategies and ground truth disinformation labels. The results reinforce previous findings, showing that across all models, \textit{Attack on Reputation}, \textit{Simplification}, \textit{Distraction}, and \textit{Manipulative Wording} exhibit positive correlations with disinformation, indicating that these strategies are strong signals of misleading content. In contrast, \textit{Justification} and \textit{Call} show in general a negligible correlation, suggesting that these strategies may be equally characteristic of credible and disinformation content. 

Table \ref{tab:matthews_correlation_detailed} extends this analysis by evaluating the correlation between persuasion strategies and final disinformation predictions under different PCoT-adapted methods (\textit{VaN}, \textit{Z-CoT}, and \textit{DeF-SpeC}). The results demonstrate consistent patterns across all configurations, suggesting that PCoT's effectiveness is not highly prompt-sensitive and remains stable across different prompting approaches.

It is important to note that we could not assess the impact of individual persuasive strategies in complete isolation, as all strategies were detected simultaneously. However, this analysis still provides valuable insight into which persuasive strategies are more characteristic of disinformation versus credible information.

\subsection{PCoT Analysis on Predicted Persuasive and Non-persuasive Content}
\label{sec:persuasiove_non_persuasive_analysis}

The results presented in Tables \ref{tab:predicted_persuasion_results_van}, \ref{tab:predicted_persuasion_results_zcot}, and \ref{tab:predicted_persuasion_results_defspec} underscore the effectiveness of the proposed Persuasion-Augmented Chain-of-Thought approach in enhancing disinformation detection across various models and prompting methods. This improvement is evident in detecting disinformation in texts with predicted persuasive strategies (\textit{Persuasion} subset) and those without (\textit{No Persuasion} subset). PCoT consistently outperforms the baseline prompting methods (\textit{VaN}, \textit{Z-CoT}, \textit{DeF-SpeC}) in the \textit{Persuasion} subset, where at least one persuasive strategy is identified. While PCoT also shows improvements in the \textit{No Persuasion} subset, the gains are lower, highlighting the challenge of detecting misleading content without persuasive cues.

\subsection{Persuasion Strategy Prediction in Disinformation and Reliable Content}
\label{sec:persuasion_detailed}

\begin{figure}[ht]  % Use a figure environment to control the placement
    \centering
    \includegraphics[width=\columnwidth]{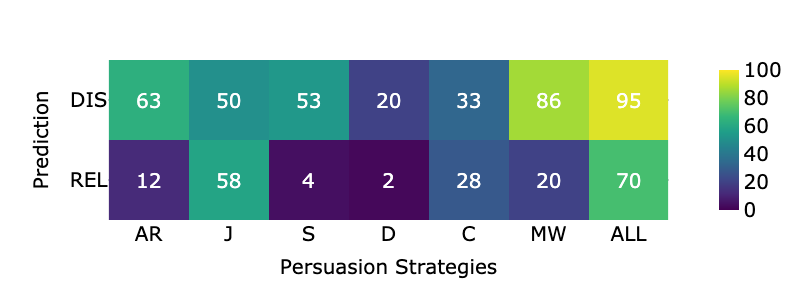}  % Replace with your image file name
    \caption{Averaged percentage of persuasion strategies predicted across 5 models in predicted disinformation (\textit{DIS}) and predicted reliable information (\textit{REL}). \textit{ALL} represents the percentage of instances with at least one detected persuasion strategy. Other abbreviations are explained in Figure \ref{fig:persuasion_strategies_tax}.}
    \label{fig:heatmap_pred}
\end{figure}

In addition to Figure \ref{fig:heatmap_true}, we provide a heatmap in Figure \ref{fig:heatmap_pred} showing the distribution of predicted persuasion strategies within the final-stage predictions of the PCoT method. Figure \ref{fig:heatmap_pred} shows that the LLM-predicted distribution of persuasion strategies for predicted disinformation and reliable information closely matches the results in Figure \ref{fig:heatmap_true}. 

\begin{figure}[ht]  % Use a figure environment to control the placement
    \centering
    \includegraphics[width=\columnwidth]{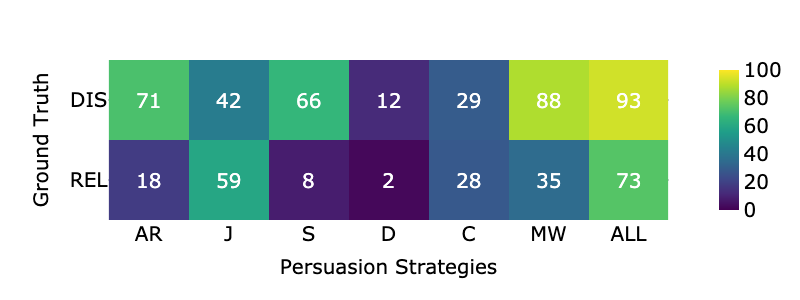}  % Replace with your image file name
    \caption{Percentage of persuasion strategies predicted by GPT 4o mini for disinformation (\textit{DIS}) and reliable information (\textit{REL}). \textit{ALL} represents the percentage of instances with at least one detected persuasion strategy. Other abbreviations are explained in Figure \ref{fig:persuasion_strategies_tax}.}
    \label{fig:gpt_4_label_heatmap}
\end{figure}

\begin{figure}[ht]  % Use a figure environment to control the placement
    \centering
    \includegraphics[width=\columnwidth]{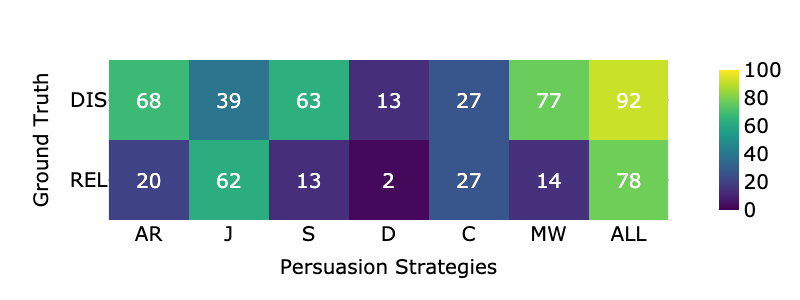}  % Replace with your image file name
    \caption{Percentage of persuasion strategies predicted by Gemini 1.5 Flash for disinformation (\textit{DIS}) and reliable information (\textit{REL}). \textit{ALL} represents the percentage of instances with at least one detected persuasion strategy. Other abbreviations are explained in Figure \ref{fig:persuasion_strategies_tax}.}
    \label{fig:gemini_label_heatmap}
\end{figure}

\begin{figure}[ht]  % Use a figure environment to control the placement
    \centering
    \includegraphics[width=\columnwidth]{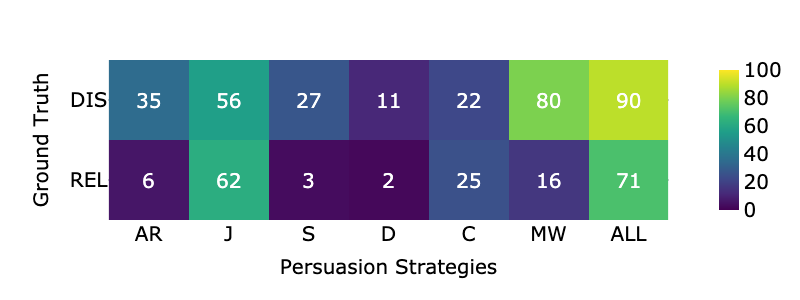}  % Replace with your image file name
    \caption{Percentage of persuasion strategies predicted by Claude 3 Haiku for disinformation (\textit{DIS}) and reliable information (\textit{REL}). \textit{ALL} represents the percentage of instances with at least one detected persuasion strategy. Other abbreviations are explained in Figure \ref{fig:persuasion_strategies_tax}.}
    \label{fig:claude_label_heatmap}
\end{figure}

\begin{figure}[ht]  % Use a figure environment to control the placement
    \centering
    \includegraphics[width=\columnwidth]{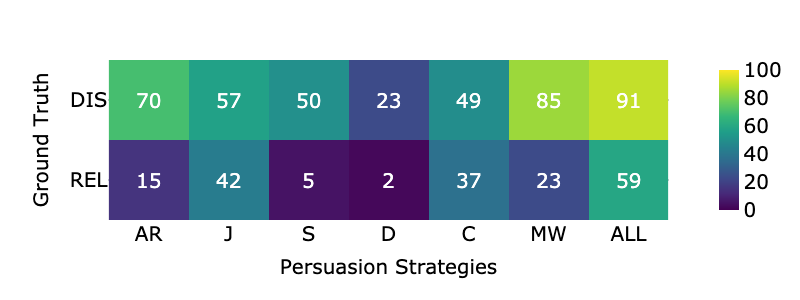}  % Replace with your image file name
    \caption{Percentage of persuasion strategies predicted by Llama 3.3 70B for disinformation (\textit{DIS}) and reliable information (\textit{REL}). \textit{ALL} represents the percentage of instances with at least one detected persuasion strategy. Other abbreviations are explained in Figure \ref{fig:persuasion_strategies_tax}.}
    \label{fig:llama70_label_heatmap}
\end{figure}

\begin{figure}[ht]  % Use a figure environment to control the placement
    \centering
    \includegraphics[width=\columnwidth]{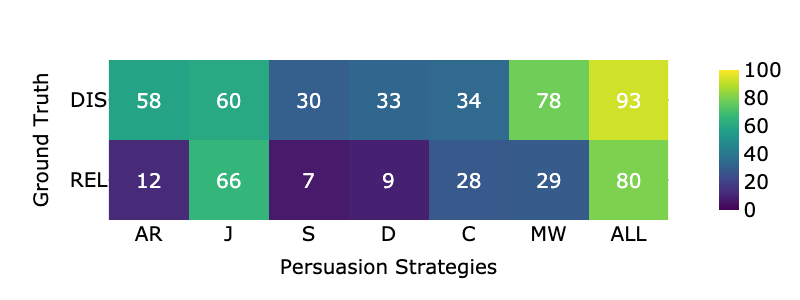}  % Replace with your image file name
    \caption{Percentage of persuasion strategies predicted by Llama 3.1 8B for disinformation (\textit{DIS}) and reliable information (\textit{REL}). \textit{ALL} represents the percentage of instances with at least one detected persuasion strategy. Other abbreviations are explained in Figure \ref{fig:persuasion_strategies_tax}.}
    \label{fig:llama8_label_heatmap}
\end{figure}

Figures \ref{fig:gpt_4_label_heatmap} to \ref{fig:llama8_label_heatmap} show heatmaps depicting the distribution of persuasion strategies across all tested models. For each model, we illustrate the distribution of predicted persuasion strategies within ground truth disinformation and reliable information. 
% Additionally, we provide heatmaps showing the distribution of predicted persuasion strategies within the final-stage predictions of the PCoT method. The general patterns observed in Figures \ref{fig:heatmap_true} and \ref{fig:heatmap_pred} are consistently reflected across experiments with individual models. 
In every case, disinformation exhibits a significantly higher percentage of texts containing at least one persuasive strategy. While credible content also employs persuasion, the distribution of specific strategies differs. \textit{Justification} and \textit{Call} are more characteristic of credible content, whereas other strategies are more commonly associated with disinformation.

\begin{table*}[ht]
\scriptsize % Make the font size smaller
\renewcommand{\arraystretch}{1.2} % Increase vertical spacing for readability
\setlength{\tabcolsep}{5pt} % Reduce horizontal spacing between columns
    \centering

\begin{tabular}{llccccc}
\toprule
Method & Data & Gemini 1.5 Flash & Claude 3 Haiku & GPT 4o mini & Llama 3.3 70B & Llama 3.1 8B\\
\midrule
VaN & overall & 0.01 & 0.01 & 0.01 & 0.01 & 0.01 \\
VaN & articles & 0.01 & 0.01 & 0.01 & 0.01 & 0.01 \\
VaN & posts & 0.01 & 0.01 & 0.01 & 0.01 & Non-Significant \\
VaN & prior & 0.01 & 0.01 & 0.01 & 0.01 & 0.01 \\
VaN & post & 0.01 & 0.01 & 0.01 & 0.01 & 0.01 \\
Z-CoT & overall & 0.01 & 0.01 & 0.01 & 0.01 & 0.01 \\
Z-CoT & articles & 0.01 & 0.01 & 0.01 & 0.01 & 0.01 \\
Z-CoT & posts & 0.01 & 0.01 & 0.01 & 0.01 & Non-Significant \\
Z-CoT & prior & 0.01 & 0.01 & 0.01 & 0.01 & 0.01 \\
Z-CoT & post & 0.01 & 0.01 & 0.01 & 0.01 & 0.01 \\
DeF-Spec & overall & 0.01 & 0.01 & 0.01 & 0.01 & 0.01 \\
DeF-Spec & articles & 0.01 & Non-Significant & 0.01 & 0.01 & 0.01 \\
DeF-Spec & posts & 0.01 & 0.01 & 0.01 & 0.01 & 0.01 \\
DeF-Spec & prior & 0.01 & Non-Significant & 0.01 & 0.01 & 0.01 \\
DeF-Spec & post & 0.01 & 0.01 & 0.01 & 0.01 & 0.05 \\
\bottomrule
\end{tabular}
     \caption{Results of McNemar's test, comparing each prompting method (\textit{VaN}, \textit{Z-CoT}, and \textit{DeF-Spec}) against its PCoT-adjusted counterpart across various language models. The values represent significance levels for different evaluation metrics, with \textit{Non-Significant} indicating no statistically significant difference at the 0.05 threshold.}
    \label{tab:mc_nemars_test}
\end{table*}

\begin{table*}[ht]
\scriptsize % Make the font size smaller
\renewcommand{\arraystretch}{1.2} % Increase vertical spacing for readability
\setlength{\tabcolsep}{5pt} % Reduce horizontal spacing between columns
    \centering

        \begin{tabular}{lrrrrrrr}
        \toprule
         & persuasion & Attack on reputation & Justification & Simplification & Distraction & Call & Manipulative wording \\
        \midrule
        GPT 4o mini & 0.228 & 0.528 & -0.160 & 0.611 & 0.230 & 0.008 & 0.507 \\
        Gemini 1.5 Flash & 0.173 & 0.476 & -0.219 & 0.511 & 0.203 & -0.000 & 0.627 \\
        Claude 3 Haiku & 0.220 & 0.378 & -0.054 & 0.354 & 0.201 & -0.029 & 0.628 \\
        Llama 3.3 70B & 0.328 & 0.546 & 0.152 & 0.536 & 0.347 & 0.118 & 0.591 \\
        Llama 3.1 8B & 0.178 & 0.484 & -0.054 & 0.301 & 0.303 & 0.064 & 0.474 \\
        \bottomrule
        \end{tabular}

    \caption{The Matthews correlation coefficient between persuasion strategies and ground truth disinformation label. Table presents coefficients for each persuasion strategy. In addition, \textit{persuasion} column shows correlation with predicted at least one persuasion strategy.}
    \label{tab:matthews_correlation}
\end{table*}

\begin{table*}[ht]
\scriptsize % Make the font size smaller
\renewcommand{\arraystretch}{1.2} % Increase vertical spacing for readability
\setlength{\tabcolsep}{5pt} % Reduce horizontal spacing between columns
    \centering
        \begin{tabular}{lrrrrrrr}
        \toprule
         & persuasion & Attack on reputation & Justification & Simplification & Distraction & Call & Manipulative wording \\
        \midrule
        \multicolumn{8}{l}{\textit{VaN with PCoT}}\\
        \hline
        GPT 4o mini & 0.307 & 0.601 & -0.187 & 0.720 & 0.279 & 0.027 & 0.592 \\
        Gemini 1.5 Flash & 0.228 & 0.495 & -0.222 & 0.638 & 0.268 & 0.036 & 0.670 \\
        Claude 3 Haiku & 0.353 & 0.491 & -0.003 & 0.466 & 0.273 & 0.011 & 0.788 \\
        Llama 3.3 70B & 0.422 & 0.648 & 0.176 & 0.622 & 0.385 & 0.158 & 0.700 \\
        Llama 3.1 8B& 0.151 & 0.479 & -0.155 & 0.362 & 0.333 & 0.027 & 0.481 \\
        \hline
        \multicolumn{8}{l}{\textit{Z-CoT with PCoT}}\\
        \hline
        GPT 4o mini & 0.308 & 0.597 & -0.183 & 0.720 & 0.273 & 0.023 & 0.585 \\
        Gemini 1.5 Flash & 0.227 & 0.495 & -0.212 & 0.640 & 0.267 & 0.034 & 0.669 \\
        Claude 3 Haiku & 0.334 & 0.504 & 0.018 & 0.419 & 0.257 & 0.012 & 0.766 \\
        Llama 3.3 70B & 0.419 & 0.642 & 0.184 & 0.625 & 0.385 & 0.154 & 0.693 \\
        Llama 3.1 8B & 0.166 & 0.484 & -0.134 & 0.356 & 0.334 & 0.026 & 0.504 \\
        \hline
        \multicolumn{8}{l}{\textit{DeF-SpeC with PCoT}}\\
        \hline
        GPT 4o mini & 0.276 & 0.558 & -0.203 & 0.720 & 0.277 & 0.025 & 0.557 \\
        Gemini 1.5 Flash & 0.250 & 0.522 & -0.225 & 0.638 & 0.260 & 0.032 & 0.709 \\
        Claude 3 Haiku & 0.346 & 0.478 & -0.003 & 0.443 & 0.255 & 0.010 & 0.782 \\
        Llama 3.3 70B & 0.395 & 0.613 & 0.159 & 0.655 & 0.408 & 0.145 & 0.667 \\
        Llama 3.1 8B & 0.163 & 0.455 & -0.161 & 0.373 & 0.340 & 0.046 & 0.477 \\
        
        \bottomrule
        \end{tabular}
    \caption{The Matthews correlation coefficient between persuasion strategies and the final 
 disinformation prediction. Table shows results for each base prompting method adopted to PCoT usage. Table presents coefficients for each persuasion strategy. In addition, \textit{persuasion} column shows correlation with predicted at least one persuasion strategy.}
    \label{tab:matthews_correlation_detailed}
\end{table*}

\begin{table}[ht]
\scriptsize
\centering
\begin{tabular}{lcccc}
\toprule
\textbf{Model} & \multicolumn{2}{c}{\textbf{Persuasion}} & \multicolumn{2}{c}{\textbf{No Persuasion}} \\
\cmidrule(lr){2-3} \cmidrule(lr){4-5}
 & \textbf{PCoT} & \textbf{Base} & \textbf{PCoT} & \textbf{Base} \\
\midrule
GPT-4o-mini & 0.876 & 0.815 & 0.315 & 0.303 \\
Gemini 1.5 Flash & 0.837 & 0.713 & 0.438 & 0.424 \\
Claude 3 Haiku & 0.840 & 0.787 & 0.128 & 0.304 \\
Llama 3.3 70B & 0.876 & 0.789 & 0.407 & 0.346 \\
Llama 3.1 8B & 0.816 & 0.631 & 0.551 & 0.561 \\
\bottomrule
\end{tabular}
\caption{Performance comparison based on F\textsubscript{1} scores across two subsets: \textit{Persuasion}, containing texts with at least one predicted persuasion strategy, and \textit{No Persuasion}, containing texts with no predicted persuasion strategies. The table reports the F\textsubscript{1} score for \textit{VaN} prompting method as \textit{Base} and for our adaptation to \textit{PCoT}.}
\label{tab:predicted_persuasion_results_van}
\end{table}

\begin{table}[ht]
\scriptsize
\centering
\begin{tabular}{lcccc}
\toprule
\textbf{Model} & \multicolumn{2}{c}{\textbf{Persuasion}} & \multicolumn{2}{c}{\textbf{No Persuasion}} \\
\cmidrule(lr){2-3} \cmidrule(lr){4-5}
 & \textbf{PCoT} & \textbf{Base} & \textbf{PCoT} & \textbf{Base} \\
\midrule
GPT-4o-mini & 0.876 & 0.827 & 0.348 & 0.297 \\
Gemini 1.5 Flash & 0.836 & 0.723 & 0.434 & 0.429 \\
Claude 3 Haiku & 0.815 & 0.644 & 0.196 & 0.206 \\
Llama 3.3 70B & 0.875 & 0.775 & 0.404 & 0.331 \\
Llama 3.1 8B & 0.818 & 0.676 & 0.535 & 0.473 \\
\bottomrule
\end{tabular}
\caption{Performance comparison based on F\textsubscript{1} scores across two subsets: \textit{Persuasion}, containing texts with at least one predicted persuasion strategy, and \textit{No Persuasion}, containing texts with no predicted persuasion strategies. The table reports the F\textsubscript{1} score for \textit{Z-CoT} prompting method as \textit{Base} and for our adaptation to \textit{PCoT}.}
\label{tab:predicted_persuasion_results_zcot}
\end{table}

\begin{table}[ht]
\scriptsize
\centering
\begin{tabular}{lcccc}
\toprule
\textbf{Model} & \multicolumn{2}{c}{\textbf{Persuasion}} & \multicolumn{2}{c}{\textbf{No Persuasion}} \\
\cmidrule(lr){2-3} \cmidrule(lr){4-5}
 & \textbf{PCoT} & \textbf{Base} & \textbf{PCoT} & \textbf{Base} \\
\midrule
GPT-4o-mini & 0.865 & 0.829 & 0.364 & 0.315 \\
Gemini 1.5 Flash & 0.861 & 0.779 & 0.459 & 0.437 \\
Claude 3 Haiku & 0.837 & 0.838 & 0.208 & 0.374 \\
Llama 3.3 70B & 0.863 & 0.780 & 0.415 & 0.351 \\
Llama 3.1 8B & 0.803 & 0.730 & 0.523 & 0.448 \\
\bottomrule
\end{tabular}
\caption{Performance comparison based on F\textsubscript{1} scores across two subsets: \textit{Persuasion}, containing texts with at least one predicted persuasion strategy, and \textit{No Persuasion}, containing texts with no predicted persuasion strategies. The table reports the F\textsubscript{1} score for \textit{DeF-SpeC} prompting method as \textit{Base} and for our adaptation to \textit{PCoT}.}
\label{tab:predicted_persuasion_results_defspec}
\end{table}

\section{Comparing BERT and LLMs on Unseen Data}
\label{sec:bert_vs_llm}

\begin{table}[!ht]
\scriptsize % Make the font size smaller
\renewcommand{\arraystretch}{1.2} % Increase vertical spacing for readability
\setlength{\tabcolsep}{5pt} % Reduce horizontal spacing between columns
    \centering
    \begin{tabular}{lll}
    \toprule
    model  & F\textsubscript{1}  Score \\
    \midrule
    BERT & 0.485 \\
    GPT 4o mini &  0.808 \\
    Gemini 1.5 Flash & 0.719 \\
    Claude 3 Haiku  & 0.677 \\
    Llama 3.3 70B &  0.752 \\
    Llama 3.1 8B  & 0.649 \\
    \bottomrule
    \end{tabular}
    \caption{Comparison between BERT performance and LLMs used with baseline methods (result averaged over 3 base methods) on post-cutoff datasets.}
    \label{tab:bert_with_base}
\end{table}

\begin{table}[!ht]
\scriptsize % Make the font size smaller
\renewcommand{\arraystretch}{1.2} % Increase vertical spacing for readability
\setlength{\tabcolsep}{5pt} % Reduce horizontal spacing between columns
    \centering
    \begin{tabular}{lll}
    \toprule
    model  & F\textsubscript{1}  Score \\
    \midrule
    BERT & 0.485 \\
    GPT 4o mini &  0.873 \\
    Gemini 1.5 Flash & 0.877 \\
    Claude 3 Haiku  & 0.783 \\
    Llama 3.3 70B &  0.864 \\
    Llama 3.1 8B  & 0.794 \\
    \bottomrule
    \end{tabular}
    \caption{Comparison between BERT performance and LLMs used with PCoT method (result averaged over 3 PCoT runs) on post-cutoff datasets.}
    \label{tab:bert_with_pcot}
\end{table}

Our experiments aim to validate the findings of \cite{lucas2023fighting}, which suggest that LLMs generalize more effectively and outperform BERT models in disinformation detection on unseen datasets. Furthermore, confirming these results strengthens the significance of our approach in advancing zero-shot classification.

\subsection{Experimental Setup}
\paragraph{Datasets} We first selected three datasets: (i) CoAID, (ii) ISOT Fake News, and (iii) ECTF, to construct our training and validation sets. The validation set contained around 6,000 texts, while the training set included approximately 40,000. For testing, we used the same subsets as in the primary PCoT evaluation experiments, enabling a direct comparison with zero-shot classification results from baseline methods and our PCoT approach. Furthermore, no articles from EUDisinfo or MultiDis were included in the training or validation sets, ensuring they remained entirely unseen by BERT.
\paragraph{Model and Optimization}
We fine-tuned widely used pre-trained BERT model. The Hugging Face model name is as follows: \texttt{google-bert/bert-large-uncased}\footnote{Hugging Face link to the BERT model and its details: \url{google-bert/bert-large-uncased}}. This model was also used by \citet{lucas2023fighting}. For our computations, including hyperparameter optimization and final fine-tuning, we utilized an NVIDIA L40 GPU. Since these experiments were not the primary focus of our study, our hyperparameter exploration was limited in scope. However, we systematically varied two key hyperparameters: learning rate and weight decay. Specifically, we experimented with learning rates ranging from 5e-6 to 5e-5 and weight decay values between 0.005 and 0.03. The final selected values were a learning rate 1e-5 and a weight decay of 0.03. Other training hyperparameters were kept constant, including a batch size of 16 for both training and evaluation, three training epochs, and a warm-up phase covering approximately 8\% of the total training steps.
\subsection{Results and Discussion}
Tables \ref{tab:bert_with_base} and \ref{tab:bert_with_pcot} present the results of our experiments comparing the baseline method and the PCoT method across various models with result on BERT model. These tables present performance of each model in detecting disinformation on unseen data, so not available during pretraining and fine-tuning of any of models. BERT performs worse than all other models, with an F\textsubscript{1} score of 0.485.

% \section{Broader Evaluation Details}
% \label{sec:broader_eval_details}

% \begin{figure}[ht]  % Use a figure environment to control the placement
%     \centering
%     \includegraphics[width=\columnwidth]{latex/figures/gpt-4o-mini_pred.png}  % Replace with your image file name
%     \caption{Averaged percentage of persuasion strategies predicted by GPT 4o mini in predicted disinformation (\textit{DIS}) and predicted reliable information (\textit{REL}) using 3 different prompting methods adapted to PCoT. \textit{ALL} represents the percentage of instances with at least one detected persuasion strategy. Other abbreviations are explained in Figure \ref{fig:persuasion_strategies_tax}.}
%     \label{fig:gpt_4_pred_heatmap}
% \end{figure}

% \begin{figure}[ht]  % Use a figure environment to control the placement
%     \centering
%     \includegraphics[width=\columnwidth]{latex/figures/gemini-1.5-flash_pred.png}  % Replace with your image file name
%     \caption{Averaged percentage of persuasion strategies predicted by Gemini 1.5 Flash in predicted disinformation (\textit{DIS}) and predicted reliable information (\textit{REL}) using 3 different prompting methods adapted to PCoT. \textit{ALL} represents the percentage of instances with at least one detected persuasion strategy. Other abbreviations are explained in Figure \ref{fig:persuasion_strategies_tax}.}
%     \label{fig:gemini_pred_heatmap}
% \end{figure}

% \begin{figure}[ht]  % Use a figure environment to control the placement
%     \centering
%     \includegraphics[width=\columnwidth]{latex/figures/claude-3-haiku-20240307_pred.png}  % Replace with your image file name
%     \caption{Averaged percentage of persuasion strategies predicted by Claude 3 Haiku in predicted disinformation (\textit{DIS}) and predicted reliable information (\textit{REL}) using 3 different prompting methods adapted to PCoT. \textit{ALL} represents the percentage of instances with at least one detected persuasion strategy. Other abbreviations are explained in Figure \ref{fig:persuasion_strategies_tax}.}
%     \label{fig:claude_pred_heatmap}
% \end{figure}

% \begin{figure}[ht]  % Use a figure environment to control the placement
%     \centering
%     \includegraphics[width=\columnwidth]{latex/figures/Llama-3.3-70B-Instruct-Turbo_pred.png}  % Replace with your image file name
%     \caption{Averaged percentage of persuasion strategies predicted by Llama 3.3 70B in predicted disinformation (\textit{DIS}) and predicted reliable information (\textit{REL}) using 3 different prompting methods adapted to PCoT. \textit{ALL} represents the percentage of instances with at least one detected persuasion strategy. Other abbreviations are explained in Figure \ref{fig:persuasion_strategies_tax}.}
%     \label{fig:llama70_pred_heatmap}
% \end{figure}

% \begin{figure}[ht]  % Use a figure environment to control the placement
%     \centering
%     \includegraphics[width=\columnwidth]{latex/figures/Meta-Llama-3.1-8B-Instruct_pred.png}  % Replace with your image file name
%     \caption{Averaged percentage of persuasion strategies predicted by Llama 3.1 8B in predicted disinformation (\textit{DIS}) and predicted reliable information (\textit{REL}) using 3 different prompting methods adapted to PCoT. \textit{ALL} represents the percentage of instances with at least one detected persuasion strategy. Other abbreviations are explained in Figure \ref{fig:persuasion_strategies_tax}.}
%     \label{fig:llama8_pred_heatmap}
% \end{figure}

\end{document}